\documentclass[10pt,twocolumn,letterpaper]{article}

\usepackage[final]{cvpr}
\usepackage[dvipsnames]{xcolor}

\usepackage{graphicx}
\usepackage{amsmath}
\usepackage{amssymb}
\usepackage{booktabs}
\usepackage{ctable}
\usepackage{multirow}
\usepackage{array}
\usepackage{colortbl}

\definecolor{cvprblue}{rgb}{0.21,0.49,0.74}
\usepackage[pagebackref,breaklinks,colorlinks,citecolor=cvprblue]{hyperref}

\def\paperID{49}
\def\confName{CVPR}
\def\confYear{2024}

\begin{document}
	
	\title{Virtually Enriched NYU Depth V2 Dataset for Monocular Depth Estimation: \\ Do We Need Artificial Augmentation?}
	
	\author{Dmitry Ignatov$^{1\thanks{Corresponding author: dmitri.ignatov@gmail.com}}$,\space\space\space\space\space\space\space\space\space\space\space\space\space\space\space\space\space\space\space\space\space\space\space\space\space Andrey Ignatov$^{2}$,\space\space\space\space\space\space\space\space\space\space\space\space\space\space\space\space\space\space\space\space\space\space\space\space\space Radu Timofte$^{1}$\\
		$^{1}$ \small{Computer Vision Lab, CAIDAS \& IFI, University of Würzburg, Germany} \\ \quad $^{2}$ \small{Computer Vision Lab, ETH Zürich, Switzerland}}
	\maketitle

	%%%%%%%%% ABSTRACT
	\begin{abstract}
		We present ANYU, a new virtually augmented version of the NYU depth v2 dataset, designed for monocular depth estimation. In contrast to the well-known approach where full 3D scenes of a virtual world are utilized to generate artificial datasets, ANYU was created by incorporating RGB-D representations of virtual reality objects into the original NYU depth v2 images. We specifically did not match each generated virtual object with an appropriate texture and a suitable location within the real-world image. Instead, an assignment of texture, location, lighting, and other rendering parameters was randomized to maximize a diversity of the training data, and to show that it is randomness that can improve the generalizing ability of a dataset. By conducting extensive experiments with our virtually modified dataset and validating on the original NYU depth v2 and iBims-1 benchmarks, we show that ANYU improves the monocular depth estimation performance and generalization of deep neural networks with considerably different architectures, especially for the current state-of-the-art VPD model. To the best of our knowledge, this is the first work that augments a real-world dataset with randomly generated virtual 3D objects for monocular depth estimation. We make our ANYU dataset publicly available in two training configurations with 10\% and 100\% additional synthetically enriched RGB-D pairs of training images, respectively, for efficient training and empirical exploration of virtual augmentation at \url{https://github.com/ABrain-One/ANYU}.
	\end{abstract}
	%%%%%%%%% BODY TEXT
	\section{Introduction}
	\label{sec:intro}
	
	This paper focuses on depth estimation from single images, a technique widely used today for 3D image synthesis in augmented reality. The concept of depth estimation refers to the determination of the distance from camera to points in a three-dimensional scene based on the analysis of two-dimensional image. The challenging process of deriving such information is key for 3D scene reconstruction and augmented reality generation, robotics and autonomous driving, essential for perception, navigation, and planning. Herewith, monocular depth estimation based on deep neural networks has demonstrated a high ability for depth prediction from RGB images~\cite{9565825, Agarwal_2023_WACV,Saxena:2023,Zhao_2023} and exhibits numerous practical applications, including the highly demanded area of mobile devices~\cite{Ignatov_2021_CVPR,10.1007/978-3-031-25066-8_4}. 
	
	To train deep depth prediction models for indoor scenes, one often uses RGB-D datasets such as NYU Depth V2~\cite{Silberman:ECCV12} (NYU-v2). The importance of the NYU-v2 is illustrated by the following statistics: according to the resource ``Papers with Code''\footnote{Dataset statistics: \url{https://paperswithcode.com/task/monocular-depth-estimation}}, the number of papers released in 2019 -- 2022 that are using indoor datasets was 439, 242, 216, and 14 for NYU-v2, Matterport3D~\cite{chang2017matterport3d}, SUN RGB-D~\cite{7298655} (a mixture of NYU-v2 with other datasets), and iBims-1 dataset~\cite{10.1016/j.cviu.2019.102877} specifically designed for validation, respectively.
	
	Despite its popularity, NYU-v2, introduced more than 10 years ago, has such drawbacks as the inaccuracy of distances seen in depth maps and limited training data diversity typical for most real-world datasets. These flaws can be partially fixed by using RGB-D images with the accurate values of depth maps from virtual reality. In contrast to the limited nature of real-world datasets, virtual reality can provide an unlimited number of 3D images with precise depth values, where objects can be captured from all possible angles and in limitless spatial combinations. In addition to the well-known approach of creating virtual datasets using complete virtual world scenes, in this paper we propose to combine 3D images from virtual reality and RGB-D data from a real-world dataset to obtain a more accurate and diverse training set of RGB-D images.
	
	\section{Related Work}
	\label{sec:formatting}
	
	Depth estimation is one of the most important tasks in the domain of scene understanding, which is often based on the training of deep models and is continuously improving with new architectures of deep convolutional neural networks and new training sets of data~\cite{MERTAN2022103441,MING202114,s20082272,chang2017matterport3d,10.1016/j.cviu.2019.102877,ipol.2023.459,10193833}. One strong trend for increasing performance in computer vision tasks is related to the use of synthetic data for training of deep convolution models~\cite{Bai_2022_CVPR,6907054,9263972,Gaidon_2016_CVPR,Ramamonjisoa_2019_ICCV,Benavides_2022_CVPR}. In particular, synthetically generated RGB-D datasets of varying complexity are utilized, ranging from simple ones, such as non-natural stacking rectangles rendering on top of each other presented by Courtois \etal~\cite{9857093}, to more advanced datasets created for indoor \cite{6907054,wu2018building,Song_2017_CVPR,McCormac_2017_ICCV} and outdoor \cite{Gaidon_2016_CVPR,Ros_2016_CVPR} scenes, where a synthetic world with high quality rendering of virtual reality details is presented.
	
	The potential effectiveness of virtual-world supervision in convolutional neural network training for monocular depth estimation was recently shown by Gurram \etal~\cite{9565825}. New synthetic RGB-D datasets provided by Roberts \etal~\cite{Roberts2021ICCV} and Zhang \etal~\cite{zhang2016physically} yield better results in some indoor depth prediction-related tasks. More sophisticated approaches, such as training on a synthetic dataset and fine-tuning on real 3D data, can improve the performance of monocular depth estimation tasks~\cite{9263972,ipol.2023.459} and enhance the occluding contours location accuracy as shown by Ramamonjisoa and Lepetit~\cite{Ramamonjisoa_2019_ICCV}.
	
	Inspired by the ideas behind the synthetic Flying Chairs dataset~\cite{ISKB18}, where 2D representations of virtual objects were successfully embedded in real-world RGB images to solve the optical flow estimation problem, in this work, we modify real RGB-D data from the NYU-v2 dataset by enriching it with randomly generated 3D virtual reality objects in order to train more accurate monocular depth estimation models.
	
	\section{Methodology of Virtual Enrichment of NYU Depth V2 Dataset}
	\label{sec:augmentation}
	
	While previous works use either only real or artificial scenes to train deep neural networks for monocular depth estimation, we consider a different approach and augment real-world data for this task with 3D objects taken from virtual reality. To obtain a higher variety of training images with more accurate depth values that can be potentially used to achieve a higher depth estimation performance, we artificially enhance real-life NYU-v2 images using virtual reality. Specifically, a new virtually augmented NYU-v2 dataset (ANYU) was generated that extends the original NYU-v2 dataset with 10\% and 100\% synthetically modified RGB-D data.

	We intentionally did not try to match each generated virtual object with the appropriate texture and place it in a suitable location. Instead, we used as much randomness as possible in generating virtual objects and choosing the place to embed them in real images to show that this way of maximizing the diversity of training data can improve the generalization ability of a dataset. The exact data generation procedure is described below. 
	
	\textbf{Input}. Virtual scenes are created using 84 3D objects of various sizes, ranging from small vases to large cabinets, and 308 specially selected diverse seamless textures that are publicly available on the Internet. For artificial modification, we randomly select RGB images and the corresponding depth maps from the entire set of 24231 pairs of RGB-D images in the NYU-v2 dataset, preserving their resolution of 640 $\times$ 480 pixels.
	
	\textbf{Virtual 3D objects generation}. Virtual 3D scene rendering is performed with a popular \textit{Unity}\footnote{\url{ https://unity.com/}} game engine~\cite{buyuksalih20173d} that is frequently used for realistic high-quality 3D object generation. For every virtual scene, we arbitrarily select number of generated objects (up to 9), parameters for their lighting, shadows, and types of virtual surfaces reflection as follows. The ``Directional'', ``Point'' or ``Spot'' light types are used equally likely. With a probability of 20\%, we choose a colored light where the brightness of red, green, and blue colors is random. From 4 to 6 light sources of the same type, with a random location behind the camera, are generated. The ``Soft'' shadows are added with a 50\% probability to a virtual object at randomly selected values of the ``shadowNormalBias'' and ``shadowBias'' parameters. The ``Standard'' or ``Diffuse'' surface reflection types are utilized for every surface separately.
	
	Seamless textures are randomly assigned to each surface of 3D objects. Affine transformation parameters are arbitrarily sampled for rendering of each virtual 3D object, its size in every dimension is scaled with a random multiplier from 0.9 to 1.1, and the colors of each texture are arbitrarily shifted in RGB space to further increase the diversity of the generated data. For each RGB-D recording, a virtual camera changes its position in the space of synthetic world to produce 3D images of virtual objects, which are subsequently utilized for augmentation of the NYU-v2 real-world RGB-D data. All the above-mentioned rendering and transformation parameters affect only virtual objects and do not directly influence the real-world scenes of the NYU-v2 dataset, which are modified only by the incorporation of virtual 3D images.
	
	\textbf{Augmentation and Culling}. We randomly select RGB-D images from the NYU-v2 training pairs and on the basis of each depth map insert RGB pixels into a Unity3D scene, calculating the share of virtual objects which remains above the surface of the original real-world image. Data pairs, where virtual objects occupy from 10\% to 50\% of the RGB image area, are used for subsequent post-processing.
	
	\textbf{Post-processing}. To maintain NYU-v2 color distribution in the virtualized RGB images, each color is normalized by the mean and standard deviation of distribution of its brightness in the original NUY-v2 training set.
	
\begin{figure*}
		\centering
		\bigskip
		\includegraphics[width=1.00\linewidth, height=10.5cm]{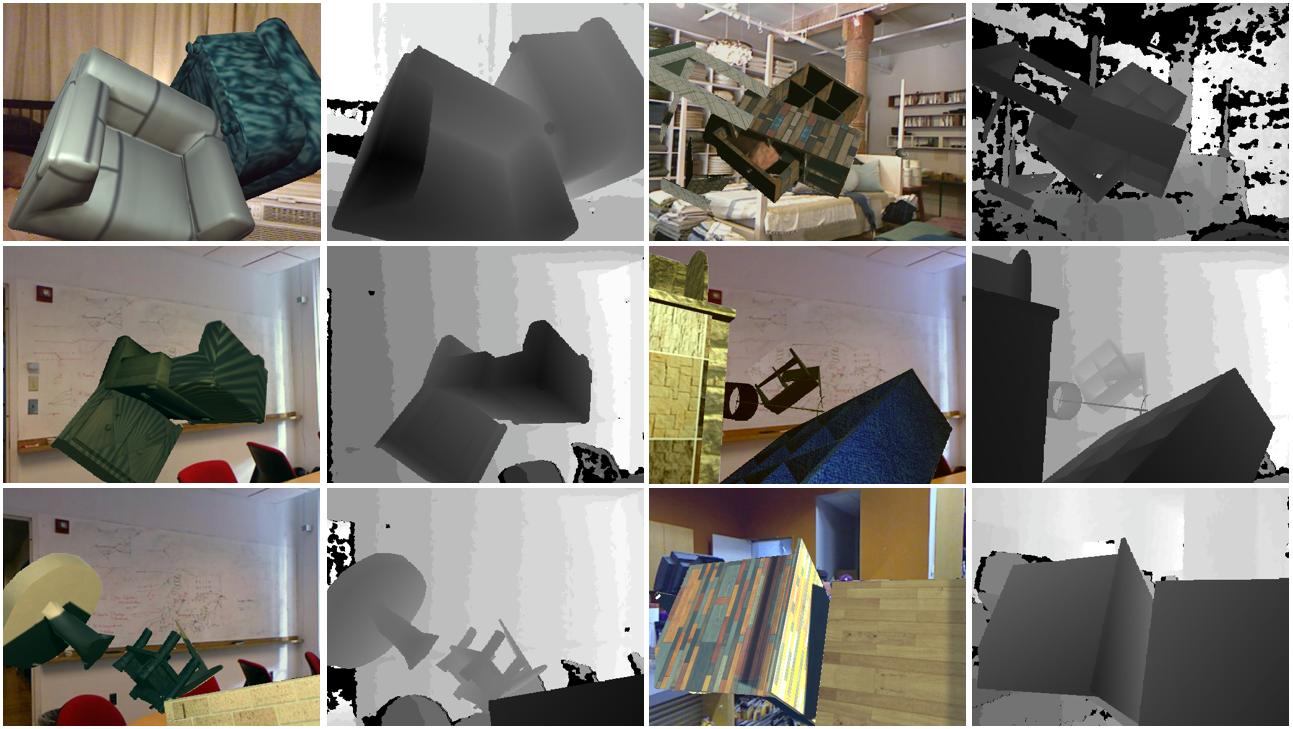}
		\includegraphics[width=1.00\linewidth, height=10.5cm]{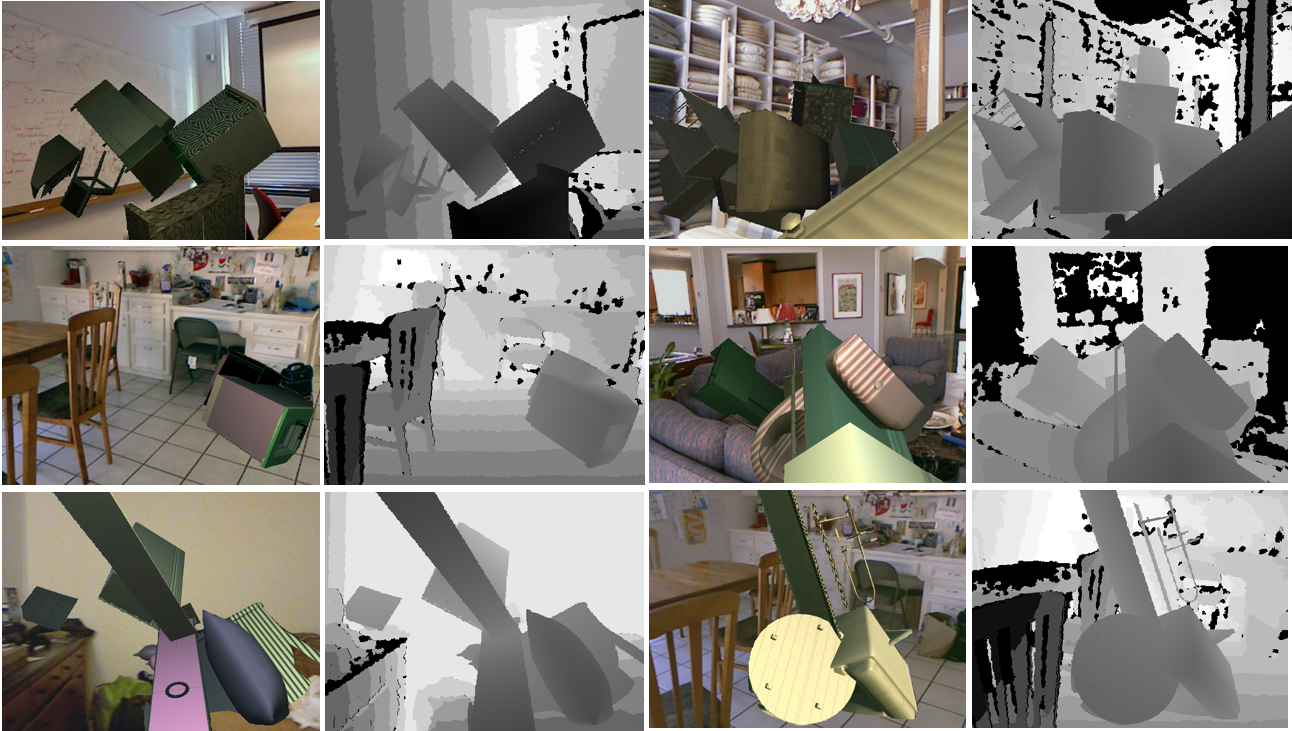}
		\caption{Examples of virtually augmented NYU-v2 RGB-D training pairs. Columns 1 and 3 show augmented RGB images, columns 2 and 4~--- the corresponding depth maps.}
		\label{rgb-d:imgs}
	\end{figure*}

	\textbf{Output}. The resulting RGB-D data are stored in the file system using the existing NYU-v2 categorization and scene naming conventions. Some examples of generated RGB-D pairs of images are shown in \cref{rgb-d:imgs}. As one can see, depth maps of virtual objects are free from random distortions, show smooth depth variation, increase visual and depth variability of images, and in such way can provide a better quality of depth estimation. 
	
	\section{Experiments}
	
	In the following sections, we empirically demonstrate the effectiveness of the augmented ANYU dataset for training of depth estimation models. For this, we train the latest state-of-the-art diffusion VPD architecture proposed by Zhao \etal~\cite{Zhao_2023} and transformer-based depth prediction PixelFormer neural network provided by Agarwal and Arora~\cite{Agarwal_2023_WACV} on the original and augmented NYU-v2 datasets. Both models trained on our ANYU dataset yield performance improvements on the NYU-v2 and iBims-1 benchmarks described below. 
	
	\subsection{Datasets}
	
	\textbf{NYU depth v2 dataset (NYU-v2)} provided by Silberman \etal~\cite{Silberman:ECCV12} contains 24231 training and 645 test images of resolution 640 $\times$ 480 pixels, and covers 464 indoor scenes. It is widely used in various research involving a fine-grained depth estimation due to the indoors origin of the captured scenes.
	
	\begin{figure}
		\includegraphics[width=1.00\linewidth]{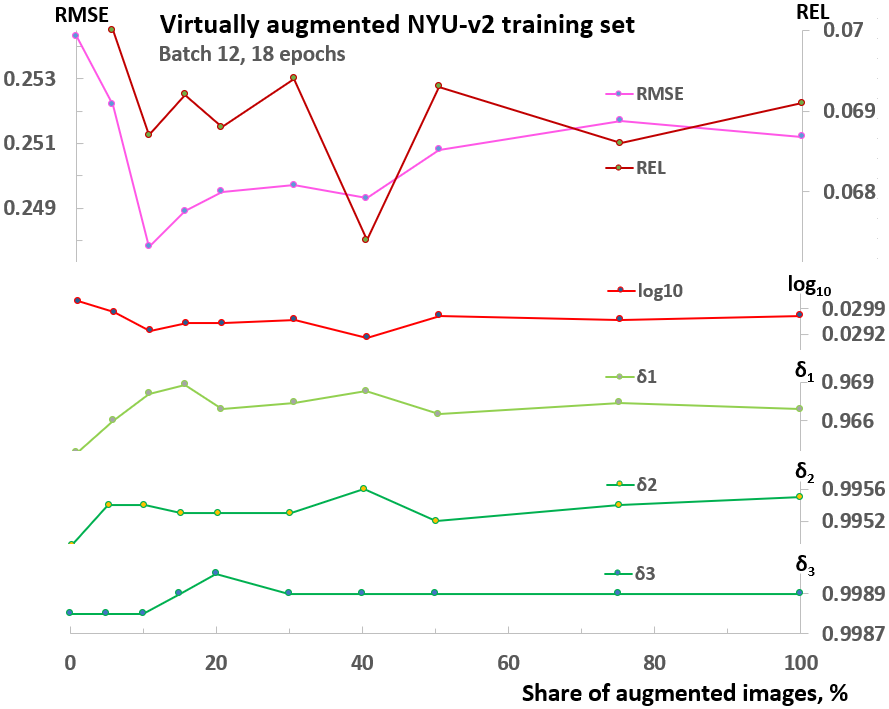}
		\caption{Performance breakdown of the VPD model~\cite{Zhao_2023} trained on the NYU-v2 dataset expanded up to a factor of 2 (100\%) with virtualized RGB-D training images. All commonly used error metrics (RMSE↓, REL↓, $\log_{10}$↓) and performance metrics ($\delta_{1}$↑, $\delta_{2}$↑, $\delta_{3}$↑) of the depth estimation show improvement over the results obtained on the original NYU-v2 dataset (0\% of augmented images, abscissa axis).}
		\label{rmse:ratio:train}	
	\end{figure}
	
	\textbf{Independent benchmark images and matched scans version 1 (iBims-1)} is employed in our experiments for cross-dataset validation. This dataset provided by Koch \etal~\cite{10.1016/j.cviu.2019.102877} consists of 100 high-quality RGB-D images and is specifically designed for validation of monocular depth estimation in different indoor scenarios. Compared to the NYU-v2, it provides lower noise levels, sharper depth transitions, fewer occlusions, and higher depth ranges. As iBims-1 has higher quality depth maps closer to real depth values compared to the NYU-v2, it can be used for more precise evaluation of depth prediction models.
	
	\subsection{Training and Testing}
	
	All experiments were performed on six NVIDIA H100 GPUs with 80GB of RAM. We used the official PyTorch implementation of the VPD model\footnote{PyTorch implementation of VPD: \url{https://github.com/wl-zhao/VPD}} provided by the authors~\cite{Zhao_2023} with a pre-trained image encoder, denoising UNet, and depth estimation decoder. The training architecture without text prompts is utilized. To reproduce the state-of-the-art results, we employed the same training and testing settings as in the original paper, except for fewer epochs and a smaller batch size, which appears to be more suitable for the virtually increased  dataset. That is why by default, VPD model was trained during 18 epochs with a batch size of 12 (6 GPUs $\times$ batch size of 2), and demonstrates the same state-of-the-art depth estimation results~(\cref{rmse:ratio:train}, 0\% of augmented images) as in the original paper~\cite{Zhao_2023}. 
	
	For the second transformer-based PixelFormer model with skip attention, we also utilized its official PyTorch implementation\footnote{PyTorch implementation of PixelFormer: \url{https://github.com/ashutosh1807/PixelFormer}} provided by Agarwal and Arora~\cite{Agarwal_2023_WACV}, preserving all training and testing parameters from this paper. In addition to the original NYU-v2 test set, we created the virtualized one according to Section~\ref{sec:augmentation}: artificially modified 2048 test RGB-D images are generated from the original 645 test pairs by enriching them with different virtual objects.
	
	\subsection{Evaluation Metrics}
	\label{sec:metrics}
	Following the standard evaluation protocol from prior works \cite{Zhao_2023,Agarwal_2023_WACV,s21010054}, for both NYU-v2 and iBims-1 datasets we report the root mean squared error (RMSE), accuracy metrics ($\delta_i < 1.25^i$ for $i \in 1, 2, 3$), absolute relative error (REL), and absolute error of log depths (log$_{10}$).
	
	\section{Results and Discussion}
	
	The models are trained by running their open-source implementations on the NYU-v2 and ANYU datasets. We evaluate the depth estimation performance with the original, virtually enriched NYU-v2 test sets, and iBims-1 validation benchmark. Subsequent sections summarize the results of experiments conducted to answer the following questions:	
	\begin{itemize}
		\item How does the accuracy on the NYU-v2 original and augmented \textit{test sets} change with the increasing percentage of artificially modified training data.
		\item What is the \textit{optimal share} of augmented training RGB-D image pairs.
		\item How \textit{cross-dataset generalization performance} of the models changes when the augmented NYU-v2 dataset is used for training.
	\end{itemize}
	
\begin{figure*}[t!]
		\centering
		\setlength{\tabcolsep}{1pt}
		\resizebox{0.93\linewidth}{!}
		{
			\begin{tabular}{ccc}
				\scriptsize{Original RGB Image}\normalsize & \scriptsize{Depth Map Generated with VPD}\normalsize & \scriptsize{Target NYU-v2 Depth Map}\normalsize\\
				\includegraphics[width=0.24\linewidth]{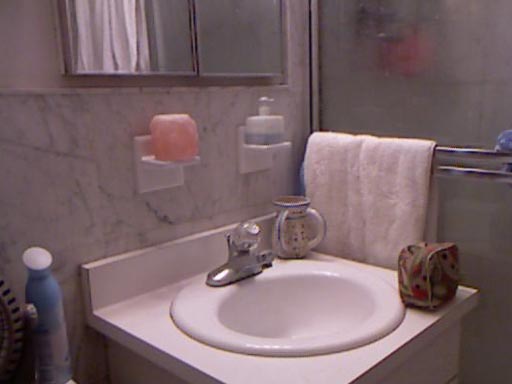}&
				\includegraphics[width=0.24\linewidth]{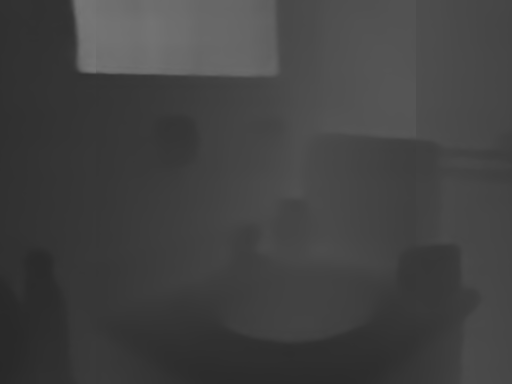}&
				\includegraphics[width=0.24\linewidth]{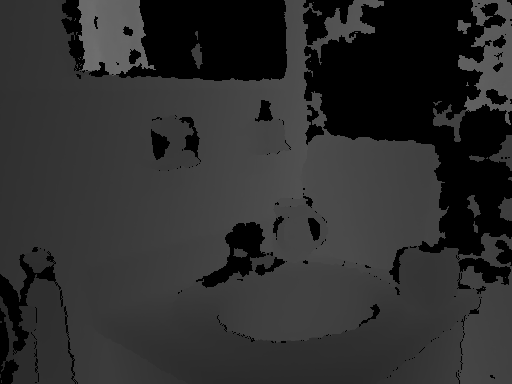}\\
				\includegraphics[width=0.24\linewidth]{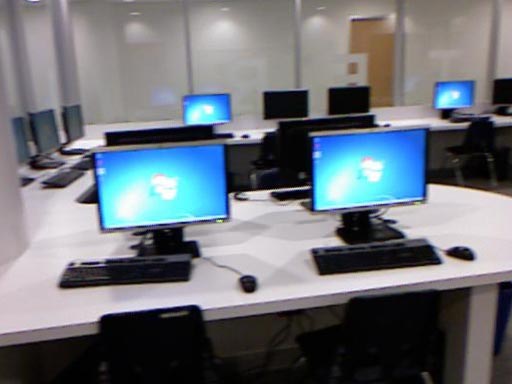}&
				\includegraphics[width=0.24\linewidth]{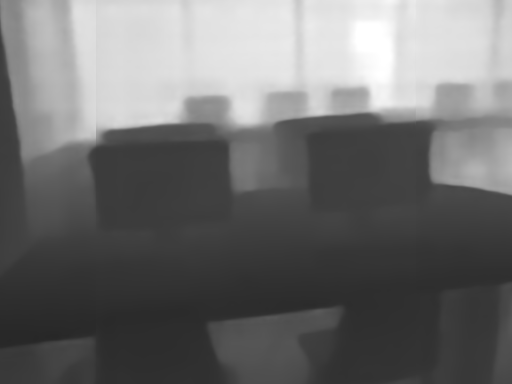}&
				\includegraphics[width=0.24\linewidth]{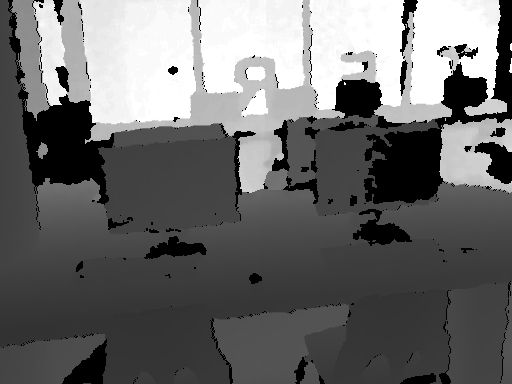}\\
				\includegraphics[width=0.24\linewidth]{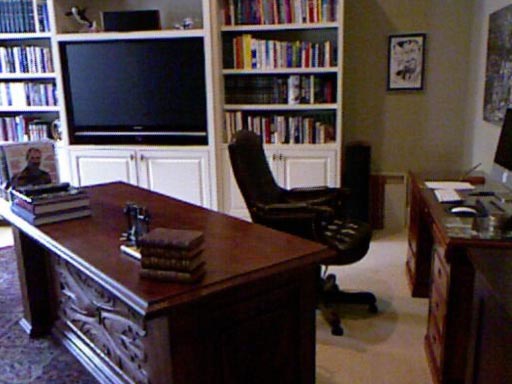}&
				\includegraphics[width=0.24\linewidth]{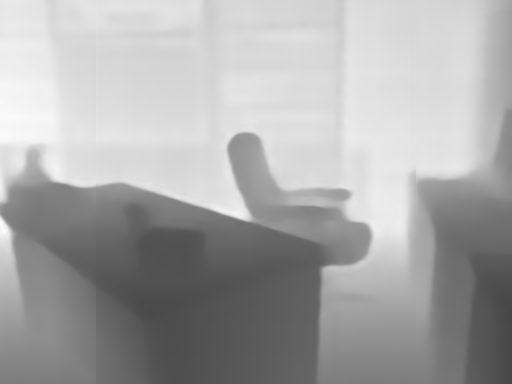}&
				\includegraphics[width=0.24\linewidth]{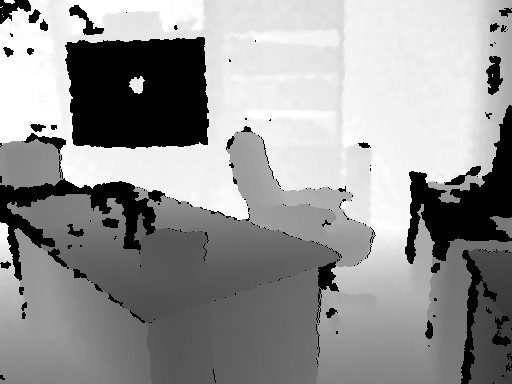}\\
				\includegraphics[width=0.24\linewidth]{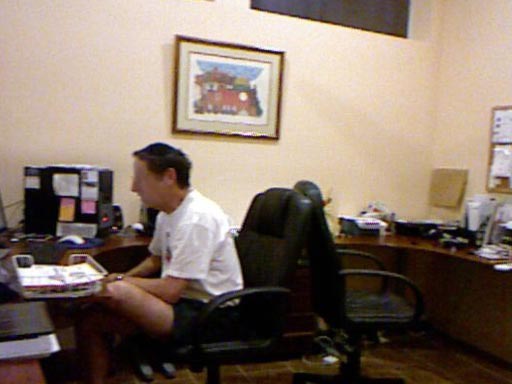}&
				\includegraphics[width=0.24\linewidth]{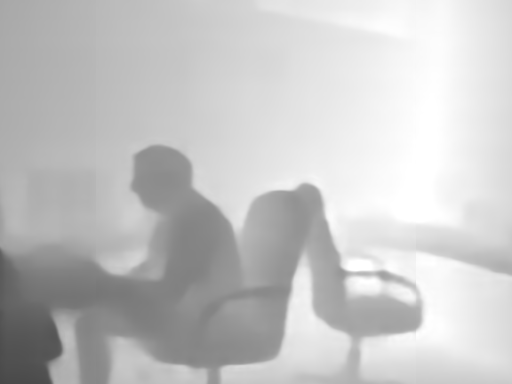}&
				\includegraphics[width=0.24\linewidth]{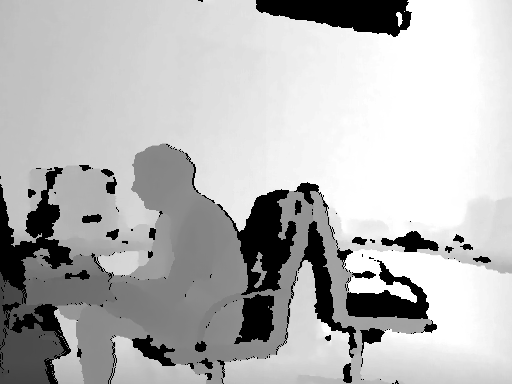}\\
				\includegraphics[width=0.24\linewidth]{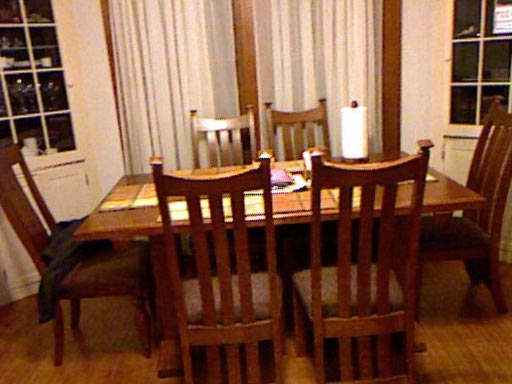}&
				\includegraphics[width=0.24\linewidth]{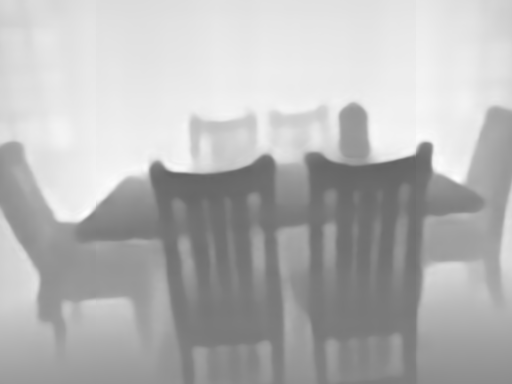}&
				\includegraphics[width=0.24\linewidth]{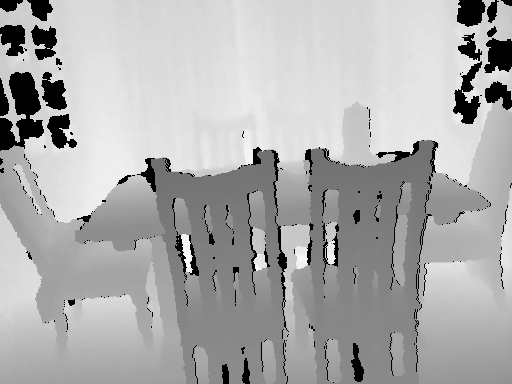}\\
			\end{tabular}
		}
		\vspace{2.0mm}
		\caption{Sample visual results obtained with the VPD model~\cite{Zhao_2023} using the proposed augmented NYU-v2 dataset (ANYU).}
		\label{fig:results}
	\end{figure*}

\begin{figure*}[t!]
		\centering
		\setlength{\tabcolsep}{1pt}
		\begin{tabular}{cccc}
			\scriptsize{Prediction w/o Augmentation}\normalsize & \scriptsize{Prediction with Augmentation}\normalsize & \scriptsize{Prediction w/o Augmentation}\normalsize & \scriptsize{Prediction with Augmentation}\normalsize\\
			\includegraphics[width=0.24\linewidth]{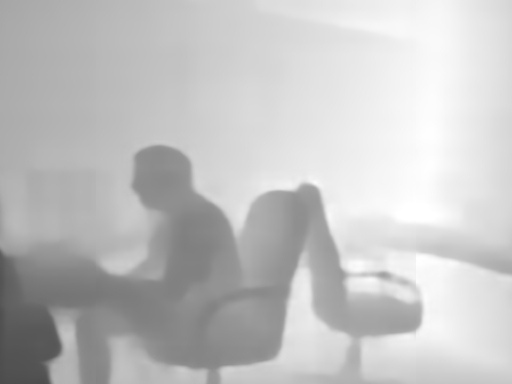}&
			\includegraphics[width=0.24\linewidth]{img/visual/study_rgb_00469-our.png}&
			\includegraphics[width=0.24\linewidth]{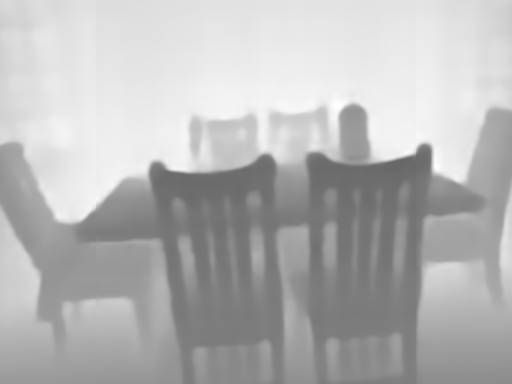}&
			\includegraphics[width=0.24\linewidth]{img/visual/dining_room_rgb_00549-our.png}\\
			\includegraphics[width=0.24\linewidth]{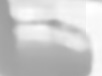}&
			\includegraphics[width=0.24\linewidth]{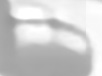}&
			\includegraphics[width=0.24\linewidth]{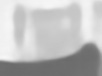}&
			\includegraphics[width=0.24\linewidth]{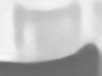}
		\end{tabular}
		\vspace{2.0mm}
		\caption{Sample visual results and the corresponding crops obtained with the VPD model~\cite{Zhao_2023} trained on the original and augmented NYU-v2 datasets. One can observe clearer, better-drawn objects and their contours when data augmentation is used.}
		\label{fig:comparison}
	\end{figure*}

	The results of these experiments are summarized in \cref{rmse:ratio:train}, \ref{rmse:ratio:test}, \ref{rmse:ratio:fraction}, \ref{rmse:ratio:reduced}, and \cref{table:rmse:nyu}, \ref{table:rmse:iBims}. Sample visual results obtained with the proposed solution are shown in \cref{fig:results}, a comparison between the results produced with and without data augmentation is illustrated in \cref{fig:comparison}. As one can see, the accuracy of depth estimation increased not only due to improved depth prediction for big 3D surfaces, but also due to better rendering of contours or even drawing of new details, which provides essentially better visual quality of 3D maps, though leads only to a slight improvement in the depth estimation metrics. Further samples of the training and test ANYU images are available in the supplementary material.
	
	\subsection{Monocular Depth Estimation on the ANYU Dataset}
	
\begin{figure}
		\includegraphics[width=1.00\linewidth]{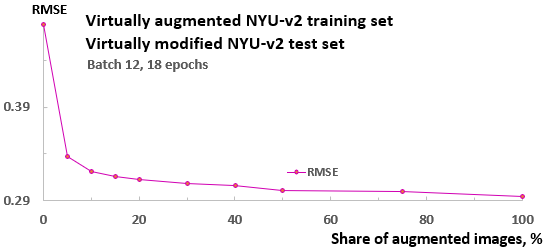}
		\caption{Performance of the VPD model \cite{Zhao_2023} tested on virtually modified NYU-v2 test set after training on the NYU-v2 dataset expanded up to a factor of 2 with virtualized RGB-D images.}
		\label{rmse:ratio:test}	
		\vspace{-3.2mm}
	\end{figure}

\begin{figure}
		\includegraphics[width=1.00\linewidth]{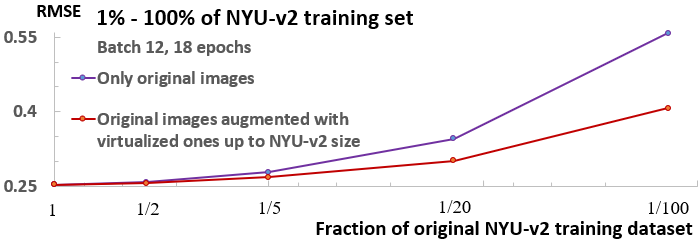}
		\caption{Performance of the VPD model \cite{Zhao_2023} trained on 1\% -- 100\% of the NYU-v2 training set. In the second series of experiments (red graph), the training set is virtually extended to the original NYU-v2 size with the proposed augmentations.}
		\label{rmse:ratio:fraction}	
	\end{figure}

\begin{figure}
		\includegraphics[width=1.00\linewidth]{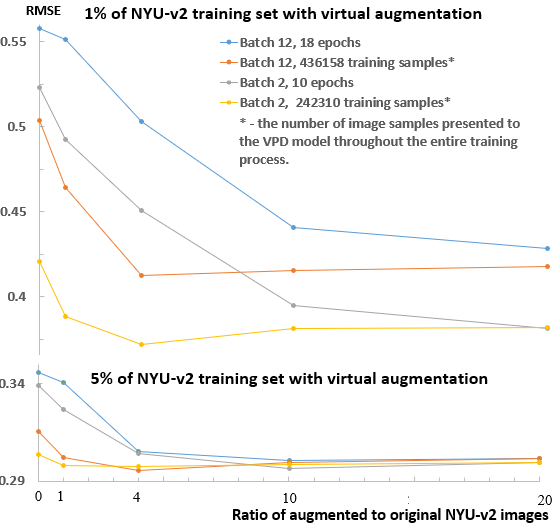}
		\caption{Performance of the VPD model \cite{Zhao_2023} trained on a small portion of the NYU-v2 dataset depending on the level of augmentation: 1\% or 5\% of the original NYU-v2 training set is expanded by up to a factor of 20 with virtualized RGB-D training images. The artificially modified dataset is utilized for training of the VPD model with a batch size of 2 or 12. A fixed number of epochs or a constant quantity of total training samples (determining the number of epochs) presented to the VPD model is used.}
		\label{rmse:ratio:reduced}
		\vspace{-2mm}
	\end{figure}

	An exhaustive quantitative comparison of typical depth estimation metrics obtained on the ANYU dataset with different proportions of the original and virtually augmented training data is presented in \cref{rmse:ratio:train}, \ref{rmse:ratio:test}, \ref{rmse:ratio:fraction}, \ref{rmse:ratio:reduced} and \cref{table:rmse:nyu}. More specifically, the results of the VPD neural network trained on RGB-D image sets with different amounts of the original and augmented training data are reported to compare their depth prediction performance on the NYU-v2 original and virtualized test sets. We quantify the augmentation-induced performance gains of computational depth prediction models in three aspects: by varying the proportion of augmented images, by reducing the NYU-v2 training set, and by altering the training conditions.
	
	\textbf{Full NYU-v2 dataset.} Depth prediction metrics for the VPD model trained on the complete NYU-v2 dataset with additional 0\%~-- 100\% virtually augmented images are summarized in statistical plots in \cref{rmse:ratio:train}. The values of the RMSE (left axis), REL, log$_{10}$, and $\delta_{1}$ -- $\delta_{3}$ (right axes) are obtained on the original NYU-v2 test set. For comparison, RMSE for depth prediction on virtually altered NYU-v2 test set is presented in~\cref{rmse:ratio:test}. As one can see, the artificial augmentation persistently effects the depth prediction for both the original~\cref{rmse:ratio:train} and the virtually modified~\cref{rmse:ratio:test} test sets. 
	
	VPD model trained on the augmented NYU-v2 dataset (5\% -- 100\%) shows lower (better) values of the RMSE, REL and log$_{10}$, and higher (better) levels of $\delta_{1}$ -- $\delta_{3}$, compared to the values of these metrics on the NYU-v2 without virtual augmentation (0\%). This means that supplementing virtually modified images yield a better depth estimation in all cases (5\% -- 100\%) according to all commonly used metrics. The lowest RMSE value is achieved when using 10\% of augmented training images, while the best values for other standard metrics are obtained at 10\% to 40\% augmentation rate (\cref{rmse:ratio:train}). Further increase in the proportion of synthetically modified training images shows a tendency to degrade the depth prediction accuracy, which, however, remains better than on the original NYU-v2 dataset without augmentation. A probable reason for this effect is an incomplete correspondence between the quality of rendering surfaces / details of virtual objects compared to real ones. This assumption is confirmed by the fact that on the virtually modified test set augmenting the training data up to 100\% leads to a continuous natural improvement in depth prediction, as shown by the constant decrease in RMSE that can be observed in~\cref{rmse:ratio:test}.
	
	In order to explore more deeply the regularities of this virtual augmentation process, we will further analyze the cases where dataset virtual enhancement yields performance gains, and in the following experiments compare the RMSE of depth prediction when reducing the size of the NYU-v2 training set.
	
	\textbf{Reduction of the NYU-v2 training set.} As one can see in \cref{rmse:ratio:fraction}, shrinking from 100\% to 1\% of the original NYU-v2 training images naturally leads to an increase in the RMSE score, while expanding the quantity of the remaining training images through their virtually enriched clones results in a substantially smaller RMSE degradation. When the number of training images is reduced down to 1\% from the initial size of the NYU-v2 training set, the RMSE rises approximately by a factor of 2, while expanding the dataset by virtually modifying the remaining training data improves the depth estimation accuracy by roughly 25\% as shown by the corresponding RMSE scores. These experiments show that, as expected, the importance of augmentation is growing with the reduction of the original training data, however, the virtually modified images cannot maintain the same quality of depth prediction as the original ones, mainly due to a lack of image background diversity.
	
	% \smallskip
	
	\textbf{Varying training conditions on 1\% and 5\% of the NYU-v2 images.} Furthermore, it is important to verify that changing the terms of VPD training, such as training parameters or a portion of supplemented images, does not impact the fact that virtual augmentation reduces the RMSE of depth prediction. Therefore, in subsequent experiments we vary the parameters of these modalities to ensure that the observed patterns of augmentation-induced improvement in depth estimation are preserved across all cases examined.
	
	As one can see from \cref{rmse:ratio:reduced}, the maximum reduction in the RMSE is observed when about 400\% of additional virtually modified RGB-D image pairs are generated from a selected fractions (1\% or 5\%) of the NYU-v2 training data. This observation affirms the intuitive understanding that increasing diversity is more important for a small number of the original images than for the entire dataset, and the smaller the image quantity, the more valuable it is to increase their diversity. For all combinations of training parameters and fractions of additional images (from 1/1 to 20/1), the RMSE value is lower than in the case when only the original images are utilized. Thus, the augmentation-induced improvement in depth prediction is preserved in all cases studied, and the positive impact of artificially modified images increases as the original image count decreases. Further details and experimental results are presented in the supplementary material.
	
	At the end of this series of experiments, we compare the impact of virtual augmentation on the performance of two fundamentally different depth prediction models.
	
	\textbf{Depth prediction models with considerably different architectures.} \cref{table:rmse:nyu} summarizes and provides qualitative comparisons of the depth prediction metrics obtained with the VPD and PixelFormer models on the original NYU-v2 (ground truth) and ANYU datasets. As one can see, along with the VPD model from a novel family of generative diffusion neural networks, the well-established transformer-based PixelFormer deep neural network also exhibits improvement for all typical depth prediction metrics, confirming that the ANYU can be used for enhancement of monocular depth estimation on different types of depth prediction deep models.
	
	\begin{table}
		\caption{Monocular depth estimation accuracy for deep models trained on the NYU depth v2 (NYU-v2) and augmented NYU-v2 (ANYU) datasets. ANYU yields new state-of-the-art results for the VPD model on the NYU-v2 test set and improves the performance of the PixelFormer on commonly accepted metrics.} \label{table:rmse:nyu}
		\centering
		\fontsize{7.5}{7.5}\selectfont
		\newcolumntype{F}{ >{\raggedright\arraybackslash} m{0.5cm} }
		\newcolumntype{G}{ >{\raggedright\arraybackslash} m{1.45cm} }
		\newcolumntype{C}{ >{\centering\arraybackslash} m{0.5cm} }
		\newcolumntype{L}{ >{\centering\arraybackslash} m{0.5cm} }
		\begin{tabular}{F G C L L L C C}
			\toprule
			Method & Dataset & RMSE↓ & $\delta_{1}$↑ & $\delta_{2}$↑ & $\delta_{3}$↑ & REL↓ & $\log_{10}$↓ \\
			\midrule 
			Pixel-Former & NYU-v2~\cite{Agarwal_2023_WACV} & 0.322 & 0.929 & 0.991 & 0.998 & 0.090 & 0.039 \\
			& \textbf{ANYU} & \textbf{0.320} & \textbf{0.930} & \textbf{0.999} & \textbf{0.999} & \textbf{0.090} & \textbf{0.038} \\  
			\midrule
			VPD &NYU-v2~\cite{Zhao_2023} & 0.254 & 0.964 & 0.995 & 0.999 &  0.069 & 0.030 \\
			& \textbf{ANYU} & \textbf{0.248} & \textbf{0.968} & \textbf{0.995} & \textbf{0.999} & \textbf{0.068} & \textbf{0.029} \\
			
			\bottomrule    
		\end{tabular}
	\end{table}
	
	The qualitative results presented in this section indicate that, in all considered scenarios, virtual augmentation leads to better performance of monocular depth estimation models compared to the training on the pure original NYU-v2. Therefore, the diversity of data in the NYU-v2 training set is not sufficient enough, and increasing it with artificial 3D images improves the results of depth estimation. 
	
	\subsection{Cross-Dataset Validation}
	
	For a decent dataset, it is important that the models trained on it can show improvement while validating on another dataset acquired from different sources, which is called cross-dataset validation. Such approach provides an assessment of generalization of prediction models, e.g., by their training on the NYU-v2 and validating on the iBims-1 dataset, as done for example by Liu \etal~\cite{s21010054}. The iBims-1 dataset contains different indoor scenarios and has higher quality depth maps closer to real depth values compared to the NYU-v2. Therefore, a validation on the iBims-1 dataset could verify the model efficiency for different data distributions between training and testing sets.
	
	We employ the same PixelFormer and VPD networks in this experiment, utilizing models trained on the NYU-v2 by their authors, and models trained in previous experiments on the ANYU (full NYU-v2 with 10\% of virtually augmented training images). These models are used without fine-tuning to generate the iBims-1 depth maps from RGB images. The common evaluation metrics for them are summarized in \cref{table:rmse:iBims}. Compared to the results~\cite{s21010054} for the depth prediction model trained on the NYU-v2 and validated on the iBims-1, the PixelForme and VPD neural networks show essential improvements in depth estimation accuracy. The virtual augmentation of the NYU-v2 increases their performance across most metrics, acknowledging on the iBims-1 validation set the generalizing ability of these ANYU-trained depth prediction models.  
	
	\begin{table}
		\caption{Cross-dataset validation on iBims-1 test set. Monocular depth estimation is performed by the models trained on the NYU depth v2 (NYU-v2) and augmented NYU-v2 (ANYU) datasets.}  \label{table:rmse:iBims}
		\centering
		\fontsize{7.5}{7.5}\selectfont
		\newcolumntype{F}{ >{\raggedright\arraybackslash} m{0.75cm} }
		\newcolumntype{G}{ >{\raggedright\arraybackslash} m{0.95cm} }
		\newcolumntype{C}{ >{\centering\arraybackslash} m{0.55cm} }
		\newcolumntype{L}{ >{\centering\arraybackslash} m{0.5cm} }
		\begin{tabular}{F G C L L L C C}
			\toprule
			Method & Training Dataset & RMSE↓ & $\delta_{1}$↑ & $\delta_{2}$↑ & $\delta_{3}$↑ & REL↓ & $\log_{10}$↓ \\
			\midrule
			Liu \etal~\cite{s21010054}  & NYU-v2 & 2.665 & 0.192 & 0.601 & 0.876 & 0.329 & 0.184 \\
			\midrule
			Pixel-Former~\cite{Agarwal_2023_WACV} & NYU-v2 & 1.595 & 0.165 & \textbf{0.553} & 0.907 & 0.335 & 0.184 \\
			& \textbf{ANYU} & \textbf{1.539} & \textbf{0.210} & 0.549 & \textbf{0.921} & \textbf{0.327} & \textbf{0.178} \\		
			\midrule		
			\textbf{VPD}~\cite{Zhao_2023} & NYU-v2 & 1.493 & 0.196 & 0.649 & \textbf{0.940} & 0.313 & 0.169 \\
			& \textbf{ANYU} & \textbf{1.365} & \textbf{0.253} & \textbf{0.677} & 0.939 & \textbf{0.294} & \textbf{0.158} \\
			\bottomrule
		\end{tabular}
	\end{table}   
	
	\section{Conclusion}
	In this paper, we introduced a new virtually augmented NYU depth v2 dataset, named ANYU, where artificially modified RGB-D training image pairs are enhanced with 3D objects from a virtual world. ANYU is provided in two training configurations with 10\% and 100\% of additional virtually enriched training images, respectively, for training new depth estimation models and for empirical exploration of the virtual augmentation. When generating ANYU, we deliberately did not match each virtual object with an appropriate texture and a suitable location within the real-world image. Instead, an assignment of texture, location, lighting, and other rendering parameters was randomized to maximize the diversity of training data, and to show that this randomness can improve the generalizability of a dataset. Considering common depth prediction metrics and validating on the NYU-v2 and iBims-1 benchmarks, we demonstrated that training depth estimation models on our ANYU data provides better generalization ability and improves the performance of indoor depth estimation. By training models with significantly different architectures on the ANYU, we improve the accuracy of the transformer-based PixelFormer model, and with the diffusion VPD neural network, achieve a new state-of-the-art result in monocular depth estimation.
	
	\section*{Acknowledgments}
	This work was partially supported by the Alexander von Humboldt Foundation.
	
	{
		\small
		\bibliographystyle{ieeenat_fullname}
		\bibliography{bibmain}
	}

\maketitlesupplementary

\begin{figure}[t!]
	\centering
	\includegraphics[width=0.49\linewidth]{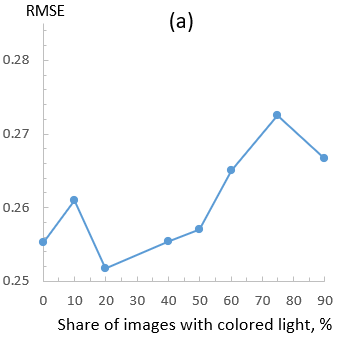}
	\includegraphics[width=0.49\linewidth]{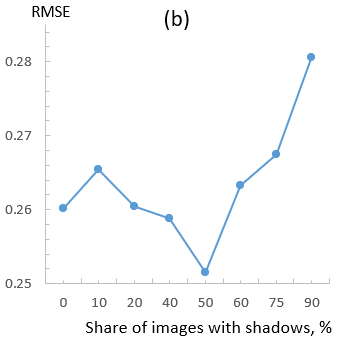}
	\caption{RMSE dependence on the share of images with colored lights (a) or with shadows (b).}
	\label{rmse:share}	
\end{figure}

Here we provide additional details that allow for a better understanding of the processes described in the main paper.

\smallskip

\section{Selecting Image Generation Parameters}
\label{image:generation}

While most of the parameters of the image virtualization process are automatically randomly selected by our Unity program over the full range of their values, some of them significantly affect the accuracy of models trained for monocular depth estimation and have to be manually chosen. These parameters include the percentage of images generated with colored light or shadows from virtual objects as illustrated in Fig.~\ref{rmse:share}. 

\smallskip

\begin{figure}[t!]
	\centering
	\includegraphics[width=0.505\linewidth]{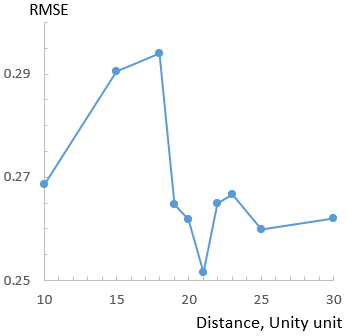}
	\caption{RMSE dependence on the distance between virtual camera and background NYU-v2 image.}
	\label{rmse:distance}	
\end{figure}

\begin{figure*}
	\centering
	\setlength{\tabcolsep}{1pt}
	\begin{tabular}{ccccc}
		\includegraphics[width=0.19\linewidth]{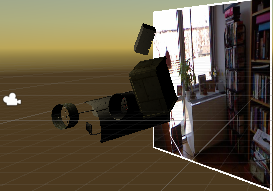}&
		\includegraphics[width=0.19\linewidth]{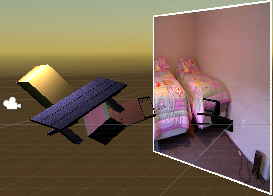}&
		\includegraphics[width=0.19\linewidth]{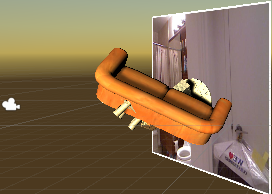}&
		\includegraphics[width=0.19\linewidth]{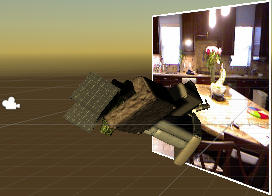}&
		\includegraphics[width=0.19\linewidth]{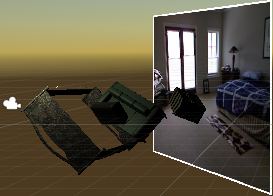}\\
		\includegraphics[width=0.19\linewidth]{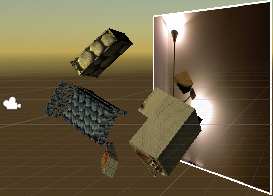}&
		\includegraphics[width=0.19\linewidth]{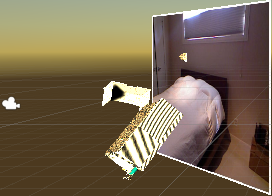}&
		\includegraphics[width=0.19\linewidth]{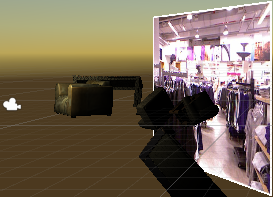}&
		\includegraphics[width=0.19\linewidth]{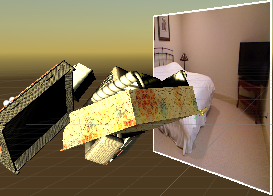}&
		\includegraphics[width=0.19\linewidth]{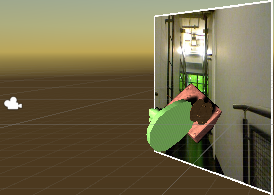}\\
		\includegraphics[width=0.19\linewidth]{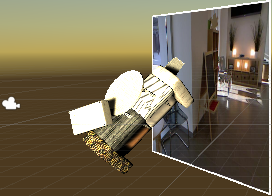}&
		\includegraphics[width=0.19\linewidth]{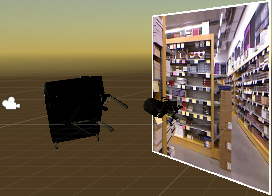}&
		\includegraphics[width=0.19\linewidth]{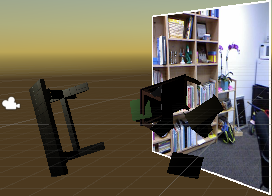}&
		\includegraphics[width=0.19\linewidth]{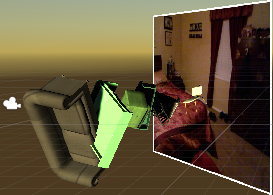}&
		\includegraphics[width=0.19\linewidth]{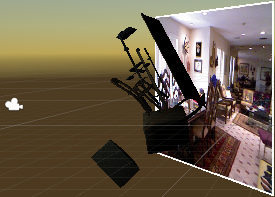}\\
		\includegraphics[width=0.19\linewidth]{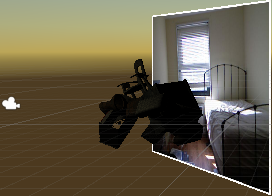}&
		\includegraphics[width=0.19\linewidth]{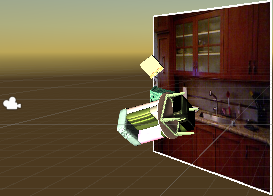}&
		\includegraphics[width=0.19\linewidth]{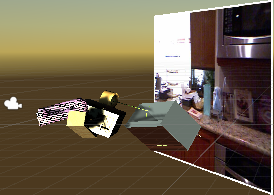}&
		\includegraphics[width=0.19\linewidth]{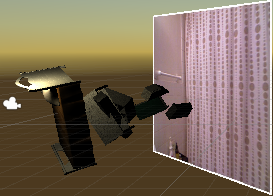}&
		\includegraphics[width=0.19\linewidth]{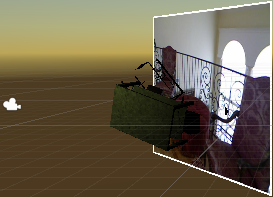}\\
		\includegraphics[width=0.19\linewidth]{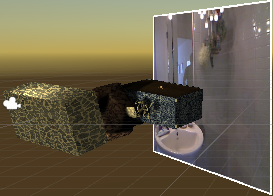}&
		\includegraphics[width=0.19\linewidth]{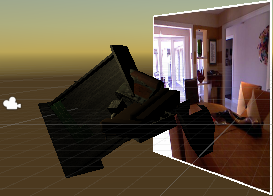}&
		\includegraphics[width=0.19\linewidth]{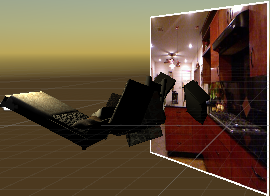}&
		\includegraphics[width=0.19\linewidth]{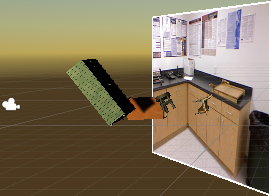}&		%		
		\includegraphics[width=0.19\linewidth]{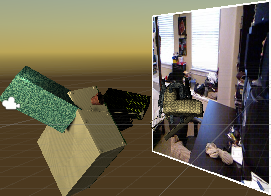}\\
		\includegraphics[width=0.19\linewidth]{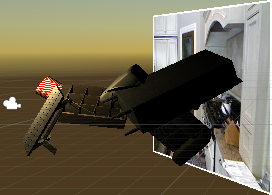}&
		\includegraphics[width=0.19\linewidth]{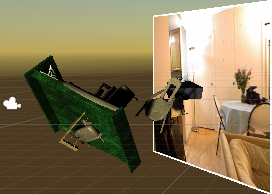}&
		\includegraphics[width=0.19\linewidth]{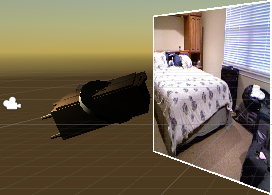}&
		\includegraphics[width=0.19\linewidth]{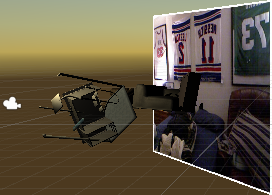}&
		\includegraphics[width=0.19\linewidth]{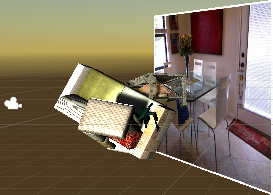}\\		%		
		\includegraphics[width=0.19\linewidth]{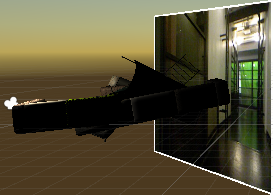}&
		\includegraphics[width=0.19\linewidth]{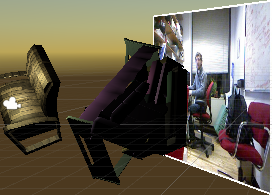}&
		\includegraphics[width=0.19\linewidth]{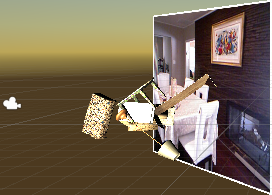}&
		\includegraphics[width=0.19\linewidth]{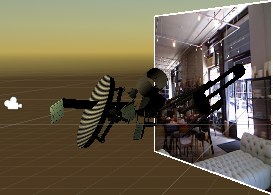}&
		\includegraphics[width=0.19\linewidth]{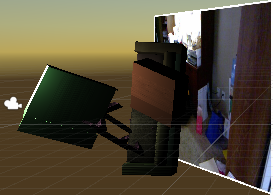}
		%&		\includegraphics[width=0.15\linewidth]{img/small_screenshots/Capture36.PNG}
	\end{tabular}
	\vspace{2.0mm}
	\caption{Screenshots of the Unity 3D at the initial stage of image virtualization: the distance between the virtual camera (left) and the NYU-v2 image plane (right) determines how many virtual objects can be effectively placed between the camera and the image plane.}
	\label{fig:small:screenshots1}
\end{figure*}

Another parameter that can be seen in \cref{fig:small:screenshots1} is the distance between virtual camera -- a small white object located near the left edge of the screenshots -- and a background image presented on the rectangular plane on the right side of the screenshots. This distance is used in the initial stage of image virtualization which is described in detail in \cref{initial:stage}. Based of the results of preliminary experiments with the VPD model presented in \cref{rmse:share} -- \ref{rmse:distance}, we set the values of the mentioned distance to 21 Unity units, and the fraction of generated images with colored lights and with shadows to 20\% and 50\%, respectively, and used these values in all subsequent experiments described in the article.

\smallskip

\section{Initial Stage of Image Virtualization}
\label{initial:stage}

\begin{figure*}[t!]
	\centering
	\setlength{\tabcolsep}{5pt}
	\begin{tabular}{c}
		\includegraphics[width=0.96\linewidth]{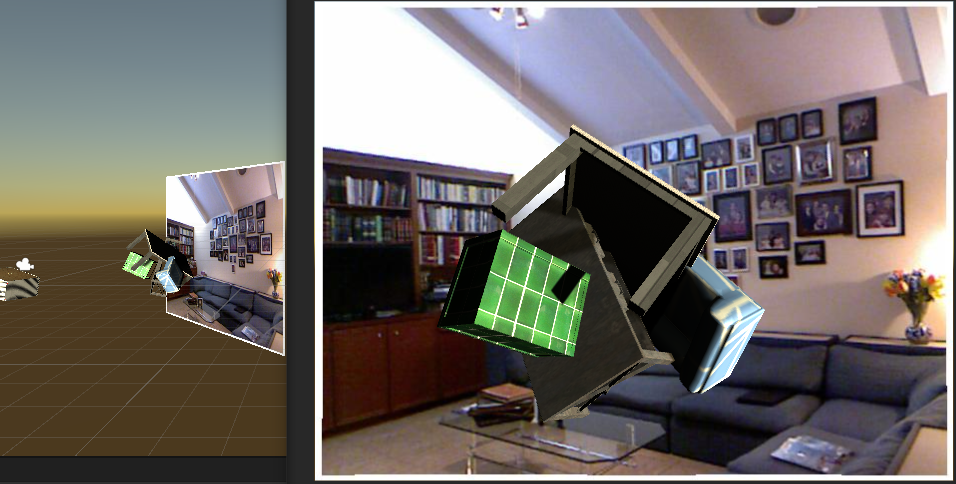}\\
		\includegraphics[width=0.96\linewidth]{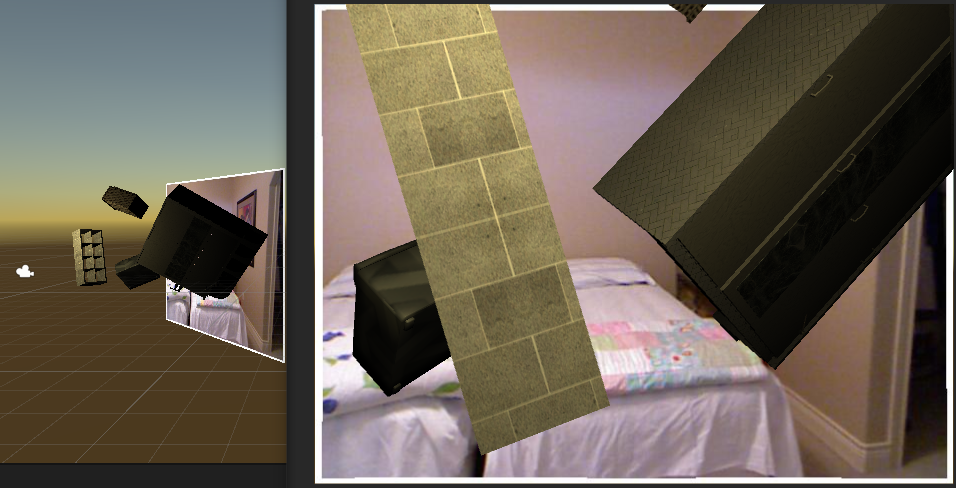}
	\end{tabular}
	\vspace{2.0mm}
	\caption{The initial stage of image virtualization in Unity 3D.}
	\label{fig:big:screenshots1}
\end{figure*}

\begin{figure*}[t!]
	\centering
	\setlength{\tabcolsep}{5pt}
	\begin{tabular}{c}
		\includegraphics[width=0.96\linewidth]{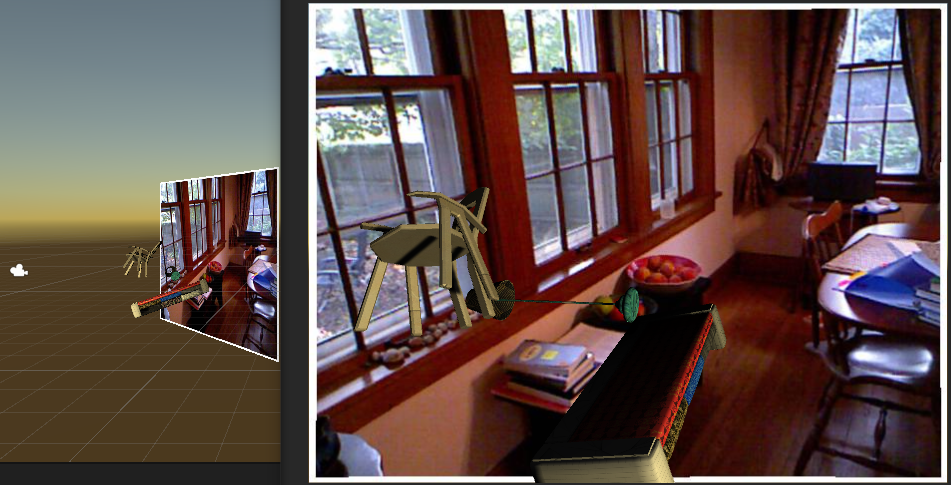}\\
		\includegraphics[width=0.96\linewidth]{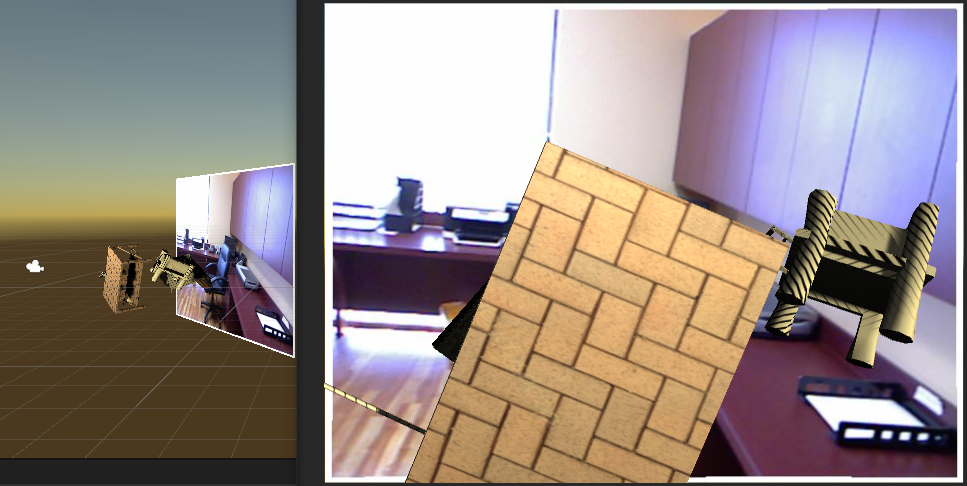}
	\end{tabular}
	\vspace{2.0mm}
	\caption{The initial stage of image virtualization in Unity 3D.}
	\label{fig:big:screenshots2}
\end{figure*}

\begin{figure*}[t!]
	\centering
	\setlength{\tabcolsep}{5pt}
	\begin{tabular}{c}
		\includegraphics[width=0.96\linewidth]{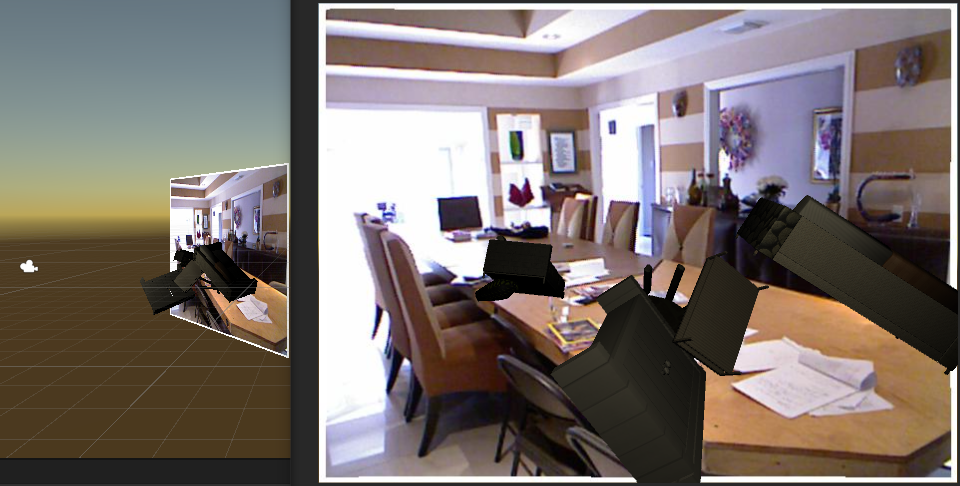}\\
		\includegraphics[width=0.96\linewidth]{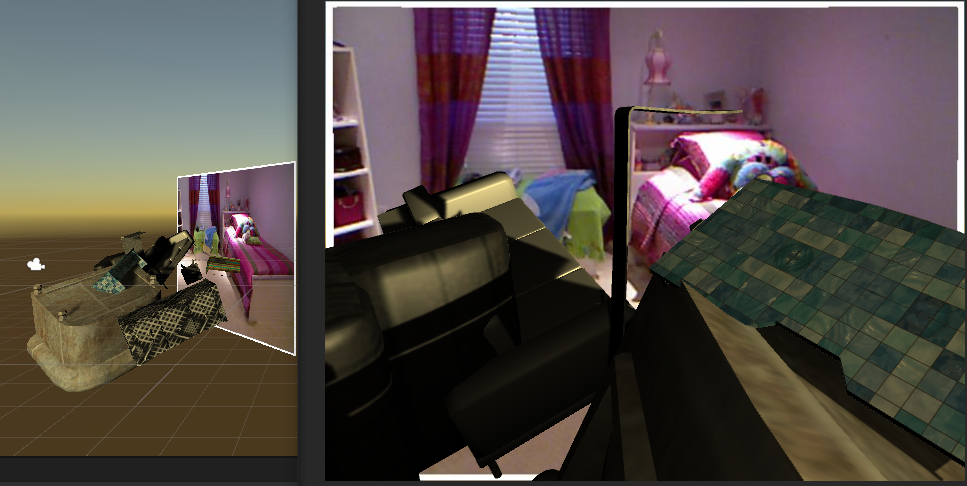}
	\end{tabular}
	\vspace{2.0mm}
	\caption{The initial stage of image virtualization in Unity 3D.}
	\label{fig:big:screenshots3}
\end{figure*}

\begin{figure*}[t!]
	\centering
	\setlength{\tabcolsep}{5pt}
	\begin{tabular}{c}		
		\includegraphics[width=0.96\linewidth]{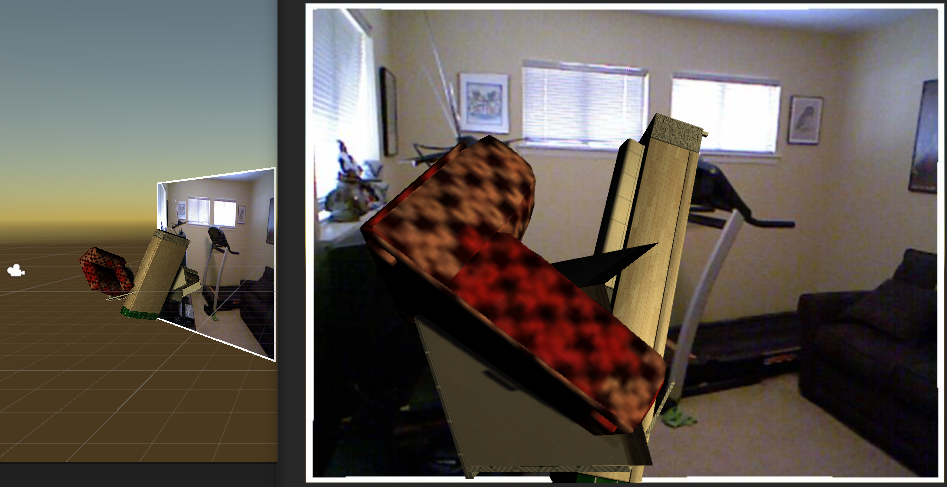}\\
		\includegraphics[width=0.96\linewidth]{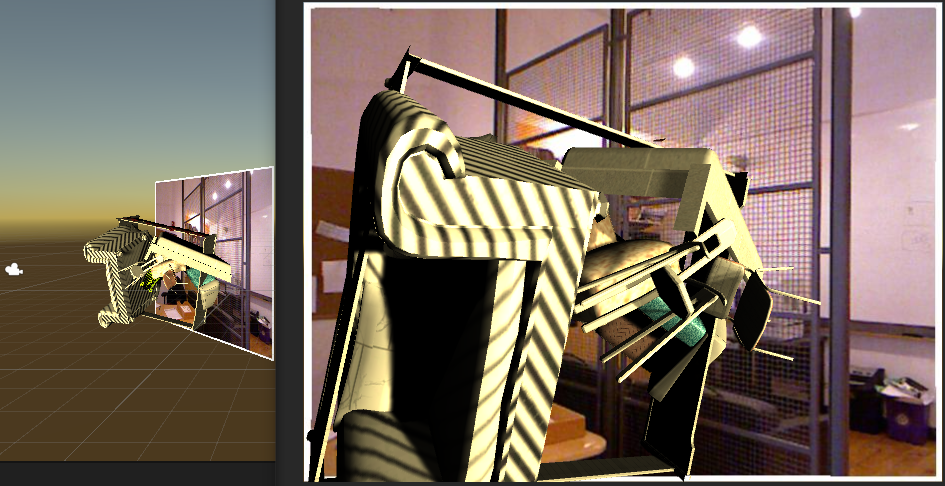}
	\end{tabular}
	\vspace{2.0mm}
	\caption{The initial stage of image virtualization in Unity 3D.}
	\label{fig:big:screenshots4}
\end{figure*}

\begin{figure*}[t!]
	\centering
	\setlength{\tabcolsep}{5pt}
	\begin{tabular}{c}
		\includegraphics[width=0.96\linewidth]{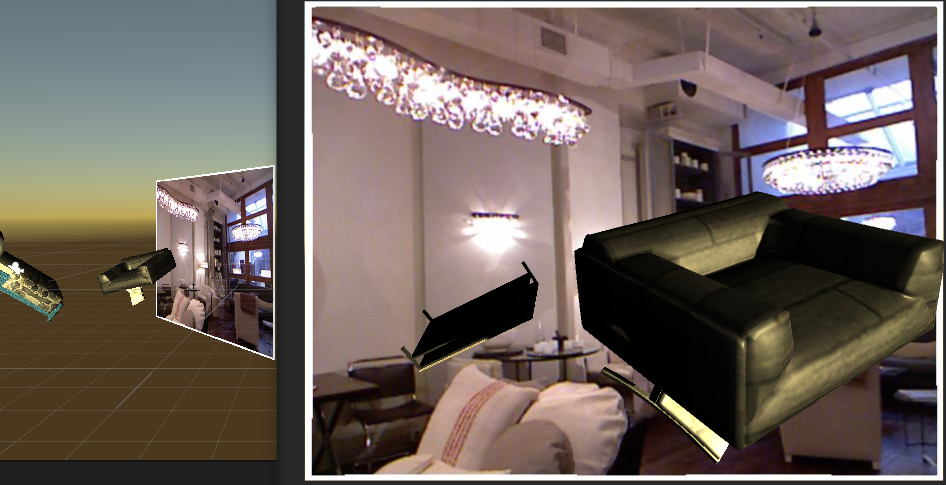} \\
		\includegraphics[width=0.96\linewidth]{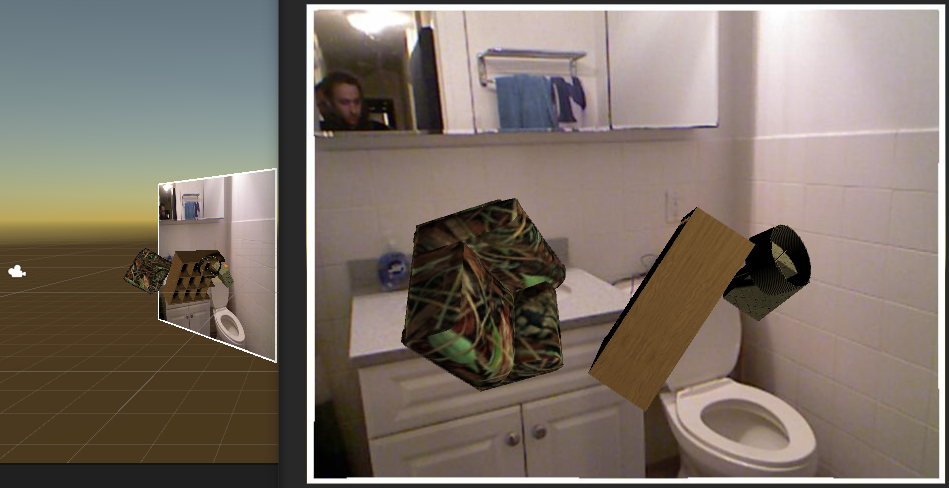}
	\end{tabular}
	\vspace{2.0mm}
	\caption{The initial stage of image virtualization in Unity 3D.}
	\label{fig:big:screenshots5}
\end{figure*}

\begin{figure*}[t!]
	\centering
	\setlength{\tabcolsep}{5pt}
	\begin{tabular}{c}
		\includegraphics[width=0.96\linewidth]{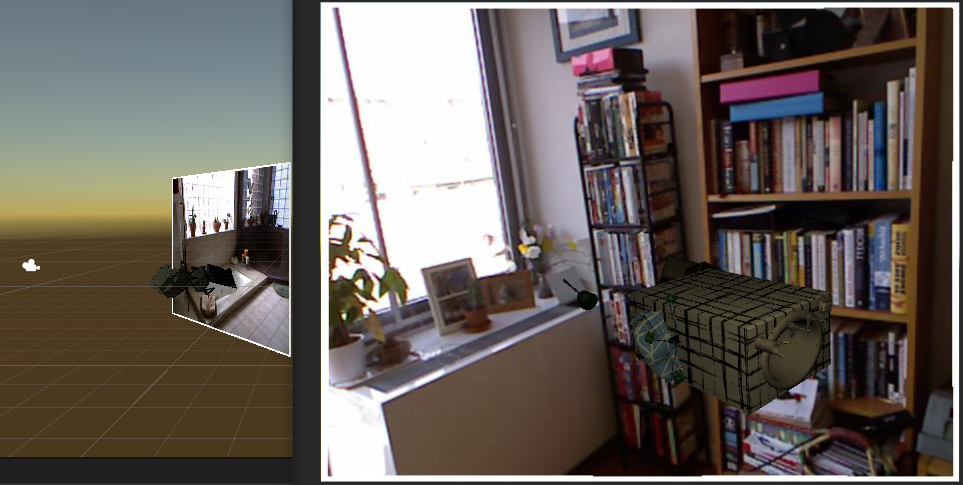}\\
		\includegraphics[width=0.96\linewidth]{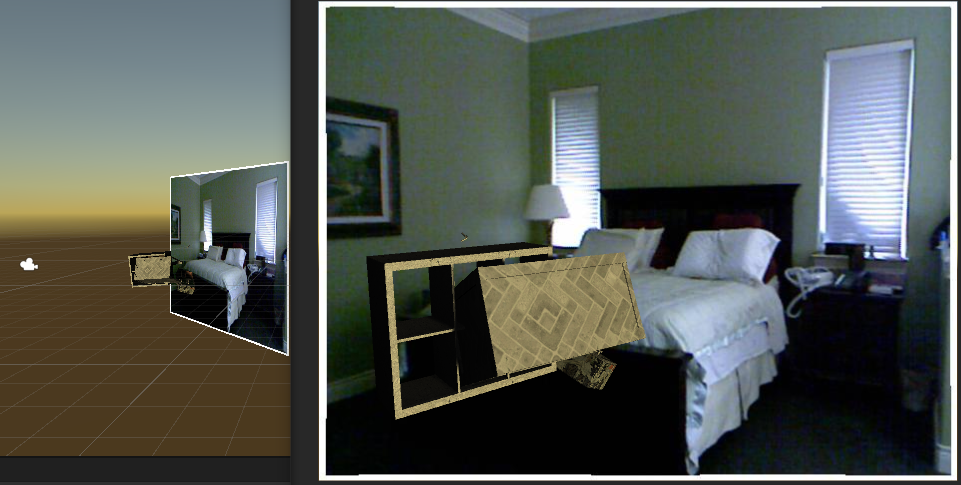}
	\end{tabular}
	\vspace{2.0mm}
	\caption{The initial stage of image virtualization in Unity 3D.}
	\label{fig:big:screenshots6}
\end{figure*}

To obtain a correct color at the edges of the virtual objects, they are rendered in the Unity 3D platform against a background of images selected from the NYU-v2 database. \cref{fig:big:screenshots1} -- \ref{fig:big:screenshots6} contain screenshots of the virtualization process of images that meet the criteria described in our paper (Sec.~\textcolor{red}{3}, \textit{Augmentation and Culling}), and therefore selected for subsequent virtualization steps. The remaining part of the screenshots represents this process in the Unity 3D space: the virtual camera on the left, virtual 3D objects in front of it, and the training image from the original NYU-v2 dataset in the background rectangular plane, which moves synchronously with the camera. On the right side of the screenshots, one can see the generated image. 

\smallskip

\section{Incorporating Virtual Objects into RGB-D Image Pairs}
\label{final:stage}

It is important to emphasize that we described above only the first step of image virtualization, where we set NYU-v2 data as the background of the image to obtain a smooth color change at the edges of the virtual objects. Afterwards, virtual objects from these images and incorporated into the original NYU-v2 RGB images, taking into account the depth maps and following the procedure described in the main paper in Sec.~\textcolor{red}{3}. After virtualization, the color distribution of new images is normalized according to the mean and standard deviation of the RGB color brightness  distribution in the original NYU-v2 training set (\cref{table:rgb}).

\begin{table}[h!]
	\caption{Color normalization parameters.} \label{table:rgb}
	\centering
	\fontsize{7.5}{7.5}\selectfont
	\newcolumntype{F}{ >{\raggedright\arraybackslash} m{0.5cm} }
	\newcolumntype{G}{ >{\raggedright\arraybackslash} m{1.5cm} }
	\newcolumntype{C}{ >{\centering\arraybackslash} m{1.5cm} }
	\newcolumntype{L}{ >{\centering\arraybackslash} m{1.5cm} }
	\begin{tabular}{F L L L }
		\toprule
		& Red & Green & Blue \\
		\midrule 
		Mean & 123.675 & 116.28 & 103.53 \\
		$\sigma$ & 58.395 & 57.12 & 57.375 \\  		
		\bottomrule    
	\end{tabular}
\end{table}

\begin{figure*}[t!]
	\centering
	\setlength{\tabcolsep}{1pt}
	\begin{tabular}{cc}
		\scriptsize{RGB image}\normalsize & \scriptsize{Depth map}\normalsize\\
		\includegraphics[width=0.49\linewidth]{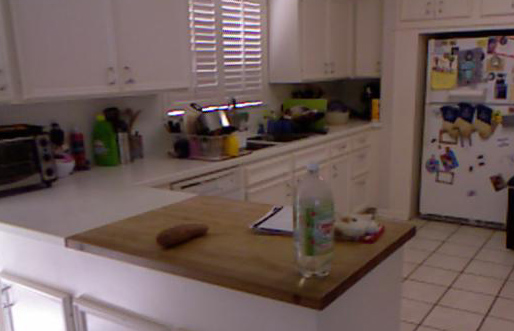}&
		\includegraphics[width=0.49\linewidth]{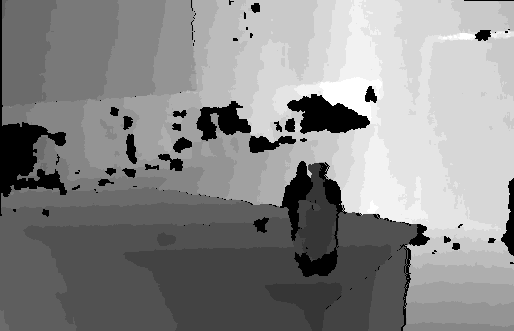}\\	
		\includegraphics[width=0.49\linewidth]{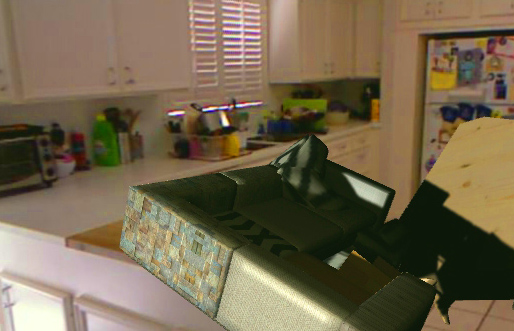}&
		\includegraphics[width=0.49\linewidth]{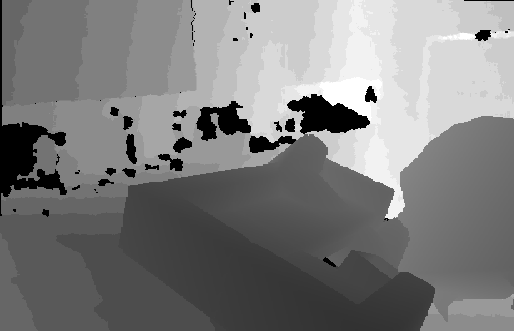}\\	\includegraphics[width=0.49\linewidth]{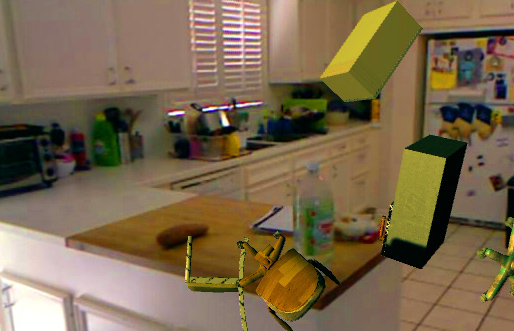}&	
		\includegraphics[width=0.49\linewidth]{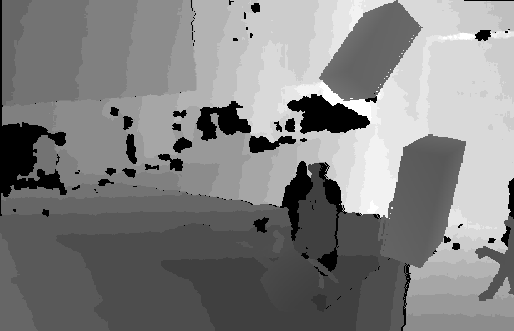}
	\end{tabular}
	\vspace{2.0mm}
	\caption{RGB-D image pairs: the original one (top) and enriched with virtual objects (the rest).}
	\label{fig:enrichment1}
\end{figure*}

\begin{figure*}[t!]
	\centering
	\setlength{\tabcolsep}{1pt}
	\begin{tabular}{cc}
		\scriptsize{RGB image}\normalsize & \scriptsize{Depth map}\normalsize\\
		\includegraphics[width=0.49\linewidth]{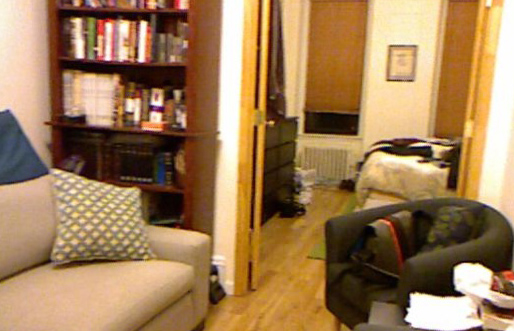}&
		\includegraphics[width=0.49\linewidth]{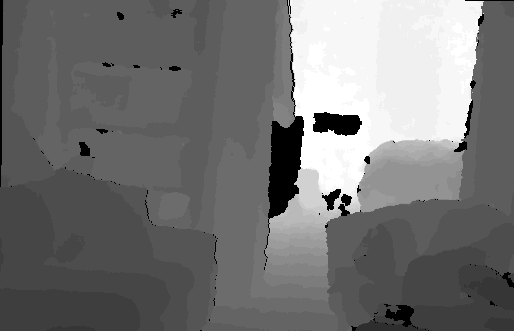}\\	
		\includegraphics[width=0.49\linewidth]{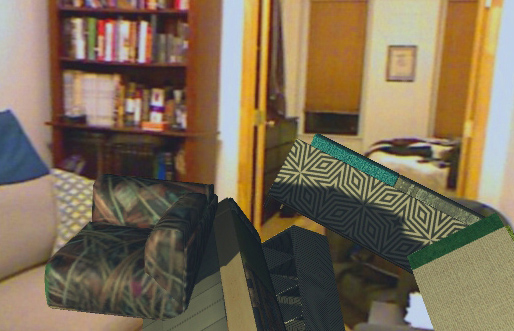}&
		\includegraphics[width=0.49\linewidth]{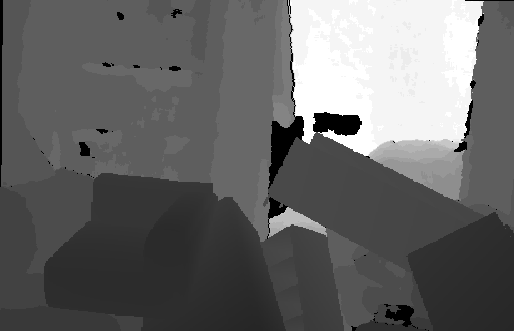}\\	\includegraphics[width=0.49\linewidth]{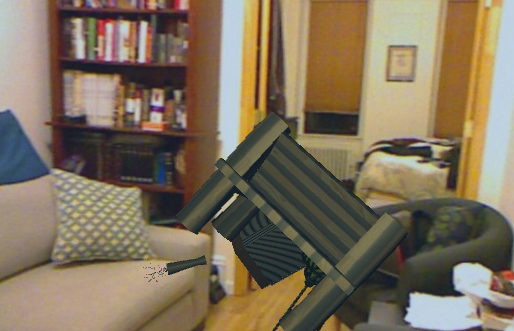}&	
		\includegraphics[width=0.49\linewidth]{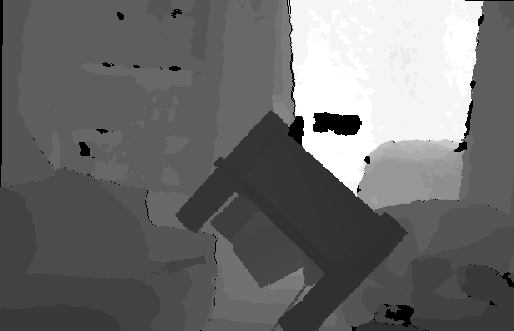}
	\end{tabular}
	\vspace{2.0mm}
	\caption{RGB-D image pairs: the original one (top) and enriched with virtual objects (the rest).}
	\label{fig:enrichment2}
\end{figure*}

\begin{figure*}[t!]
	\centering
	\setlength{\tabcolsep}{1pt}
	\begin{tabular}{cc}
		\scriptsize{RGB image}\normalsize & \scriptsize{Depth map}\normalsize\\
		\includegraphics[width=0.49\linewidth]{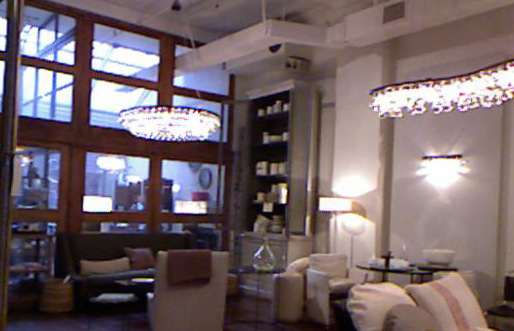}&
		\includegraphics[width=0.49\linewidth]{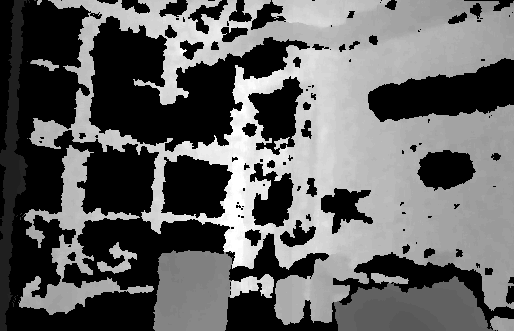}\\	
		\includegraphics[width=0.49\linewidth]{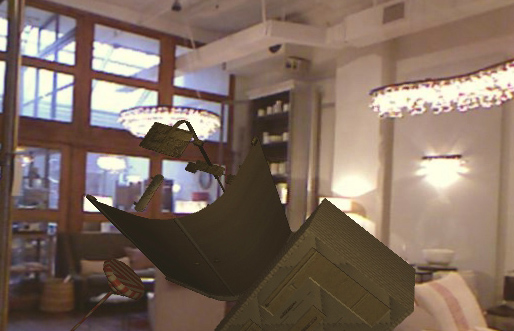}&
		\includegraphics[width=0.49\linewidth]{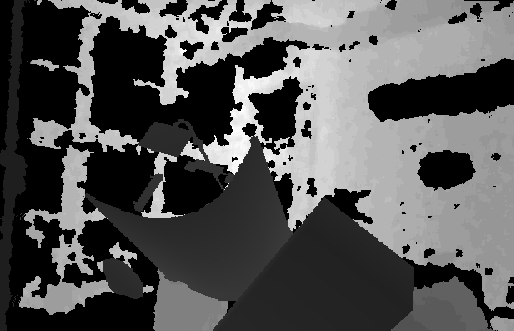}\\	\includegraphics[width=0.49\linewidth]{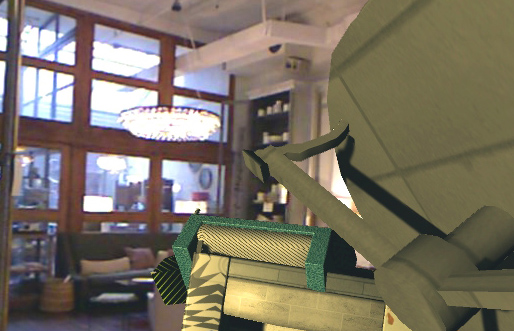}&	
		\includegraphics[width=0.49\linewidth]{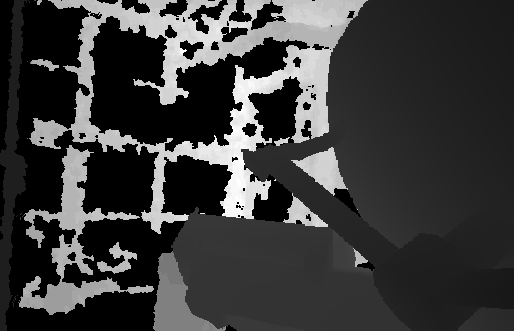}
	\end{tabular}
	\vspace{2.0mm}
	\caption{RGB-D image pairs: the original one (top) and enriched with virtual objects (the rest).}
	\label{fig:enrichment3}
\end{figure*}

\begin{figure*}[t!]
	\centering
	\setlength{\tabcolsep}{1pt}
	\begin{tabular}{cc}
		\scriptsize{RGB image}\normalsize & \scriptsize{Depth map}\normalsize\\
		\includegraphics[width=0.49\linewidth]{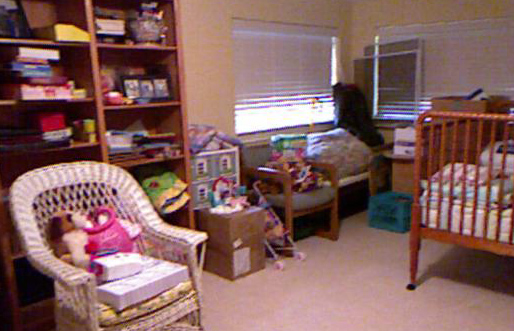}&
		\includegraphics[width=0.49\linewidth]{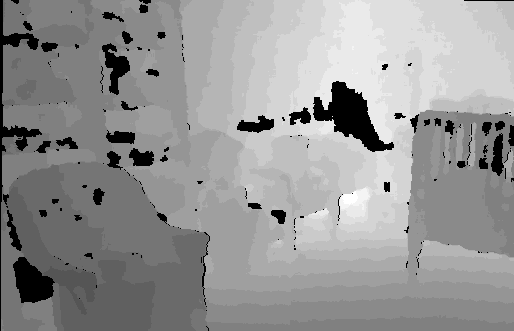}\\	
		\includegraphics[width=0.49\linewidth]{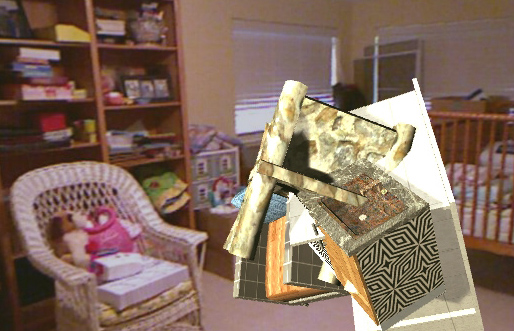}&
		\includegraphics[width=0.49\linewidth]{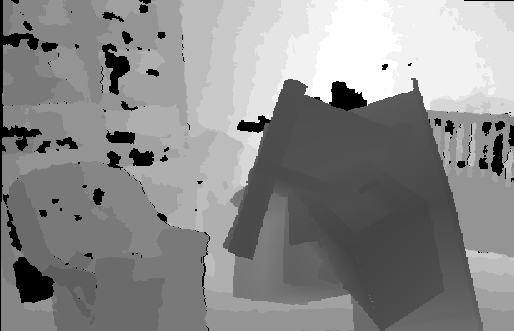}\\	\includegraphics[width=0.49\linewidth]{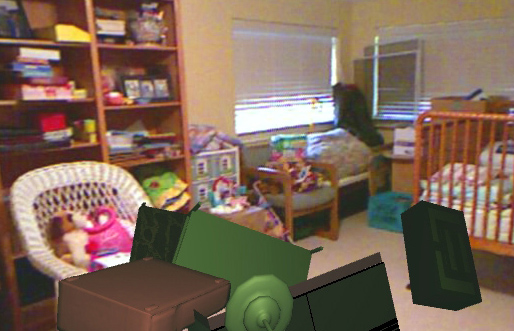}&	
		\includegraphics[width=0.49\linewidth]{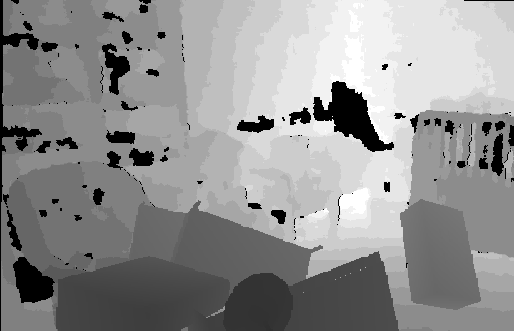}
	\end{tabular}
	\vspace{2.0mm}
	\caption{RGB-D image pairs: the original one (top) and enriched with virtual objects (the rest).}
	\label{fig:enrichment4}
\end{figure*}

\begin{figure*}[t!]
	\centering
	\setlength{\tabcolsep}{1pt}
	\begin{tabular}{cc}
		\scriptsize{RGB image}\normalsize & \scriptsize{Depth map}\normalsize\\
		\includegraphics[width=0.49\linewidth]{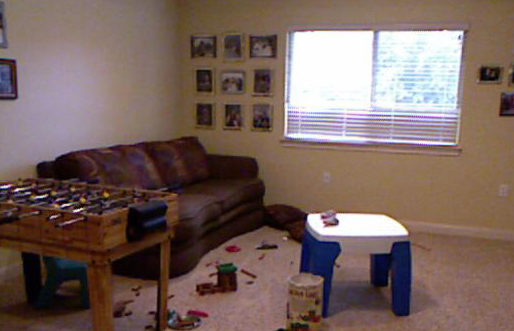}&
		\includegraphics[width=0.49\linewidth]{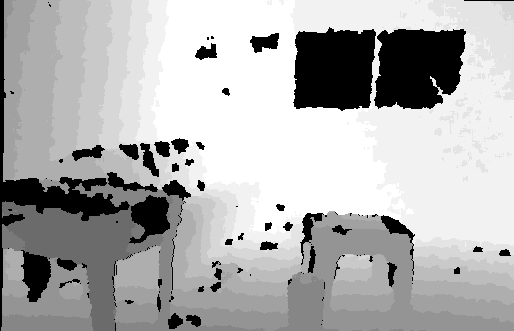}\\	
		\includegraphics[width=0.49\linewidth]{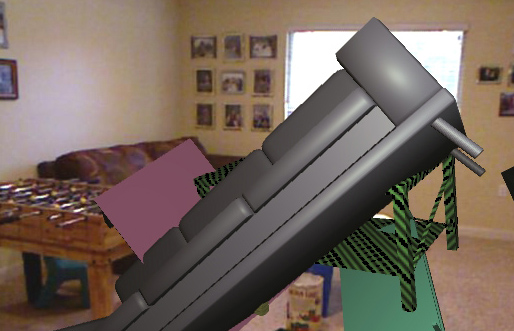}&
		\includegraphics[width=0.49\linewidth]{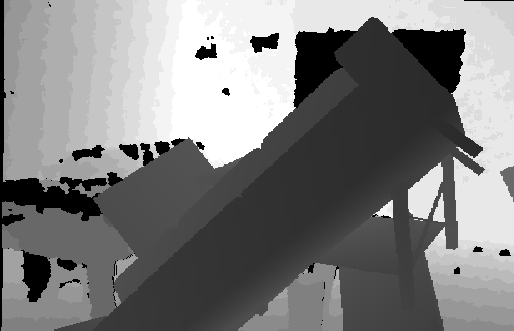}\\	\includegraphics[width=0.49\linewidth]{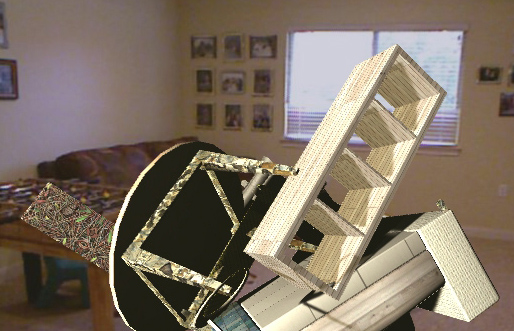}&	
		\includegraphics[width=0.49\linewidth]{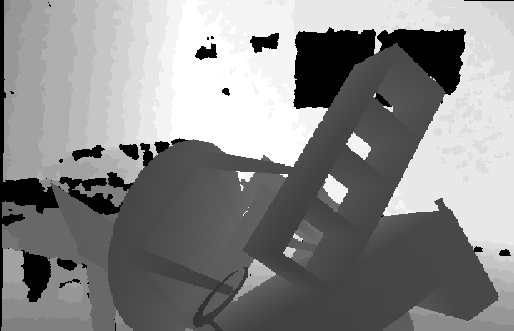}
	\end{tabular}
	\vspace{2.0mm}
	\caption{RGB-D image pairs: the original one (top) and enriched with virtual objects (the rest).}
	\label{fig:enrichment5}
\end{figure*}

\begin{figure*}[t!]
	\centering
	\setlength{\tabcolsep}{1pt}
	\begin{tabular}{cc}
		\scriptsize{RGB image}\normalsize & \scriptsize{Depth map}\normalsize\\
		\includegraphics[width=0.49\linewidth]{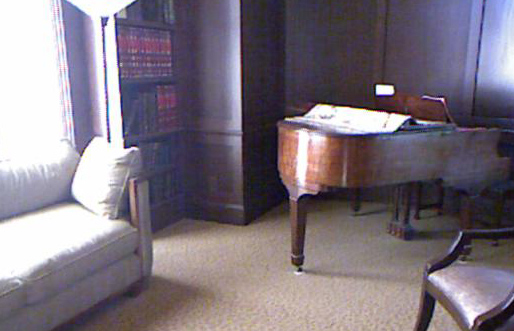}&
		\includegraphics[width=0.49\linewidth]{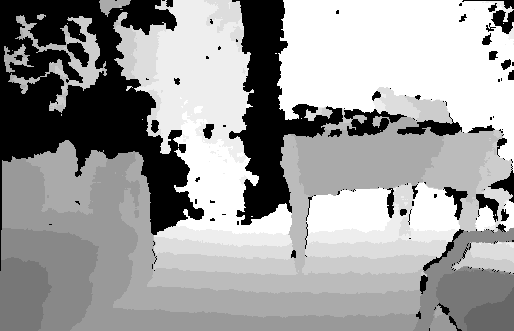}\\	
		\includegraphics[width=0.49\linewidth]{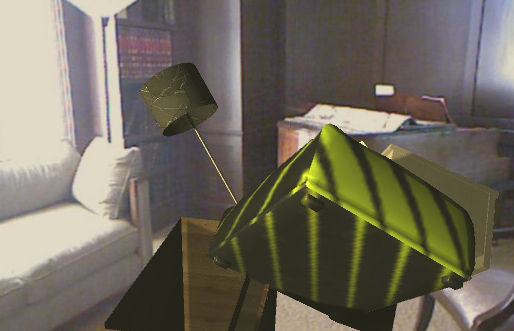}&
		\includegraphics[width=0.49\linewidth]{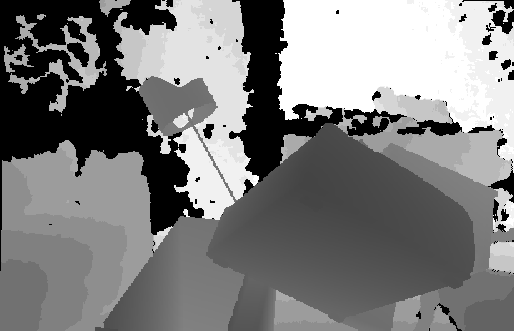}\\	\includegraphics[width=0.49\linewidth]{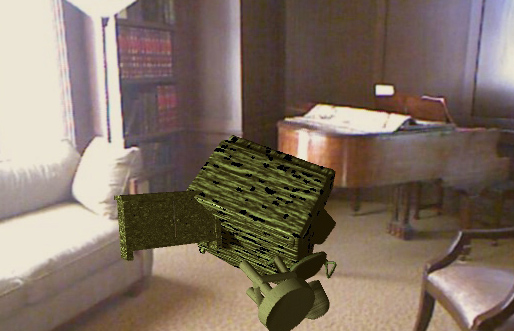}&	
		\includegraphics[width=0.49\linewidth]{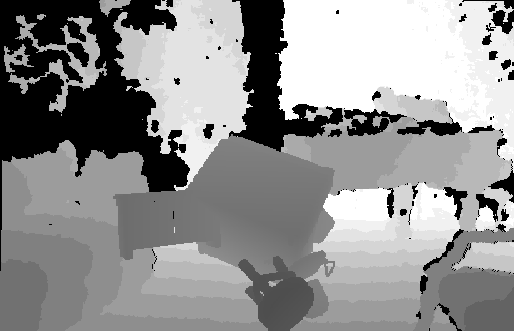}
	\end{tabular}
	\vspace{2.0mm}
	\caption{RGB-D image pairs: baseline (first from top) and enriched with virtual objects (the rest).}
	\label{fig:enrichment6}
\end{figure*}

\begin{figure*}[t!]
	\centering
	\setlength{\tabcolsep}{1pt}
	\begin{tabular}{cc}
		\scriptsize{RGB image}\normalsize & \scriptsize{Depth map}\normalsize\\
		\includegraphics[width=0.49\linewidth]{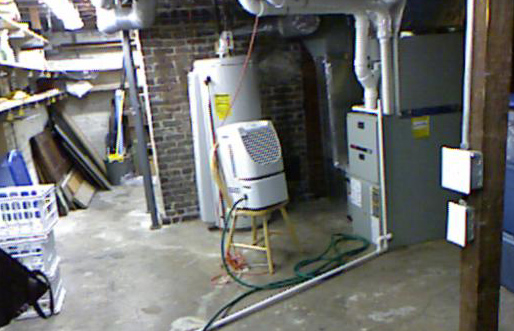}&
		\includegraphics[width=0.49\linewidth]{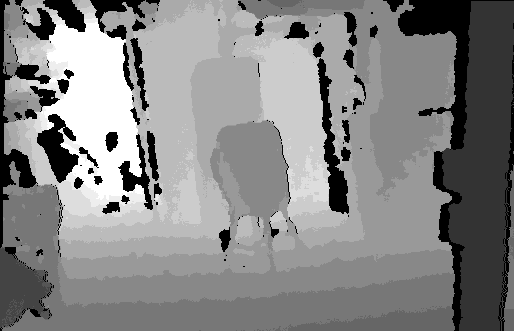}\\	
		\includegraphics[width=0.49\linewidth]{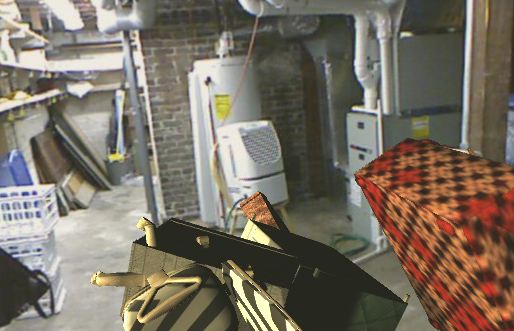}&	
		\includegraphics[width=0.49\linewidth]{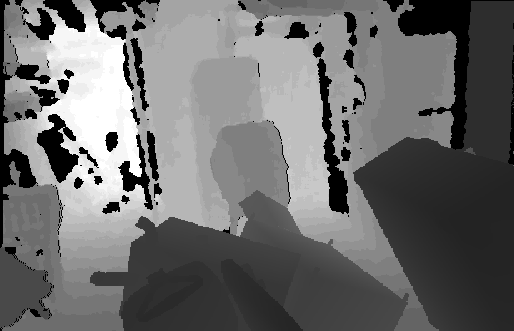}\\
		\includegraphics[width=0.49\linewidth]{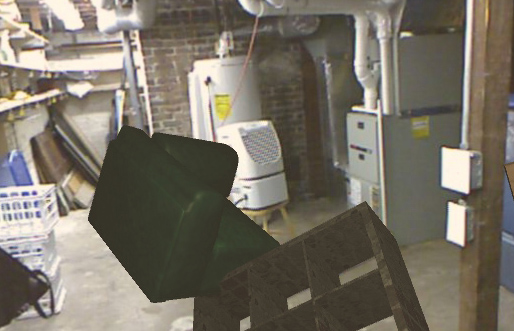}&
		\includegraphics[width=0.49\linewidth]{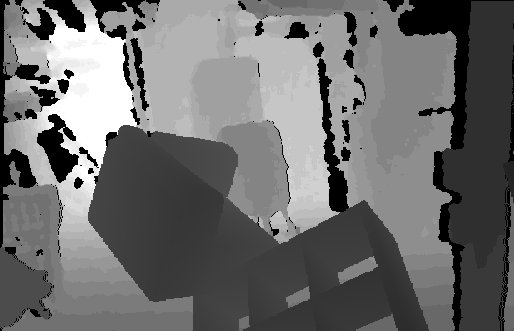}	
	\end{tabular}
	\vspace{2.0mm}
	\caption{RGB-D image pairs: the original one (top) and enriched with virtual objects (the rest).}
	\label{fig:enrichment7}
\end{figure*}

\begin{figure*}[t!]
	\centering
	\setlength{\tabcolsep}{1pt}
	\begin{tabular}{cc}
		\scriptsize{RGB image}\normalsize & \scriptsize{Depth map}\normalsize\\
		\includegraphics[width=0.49\linewidth]{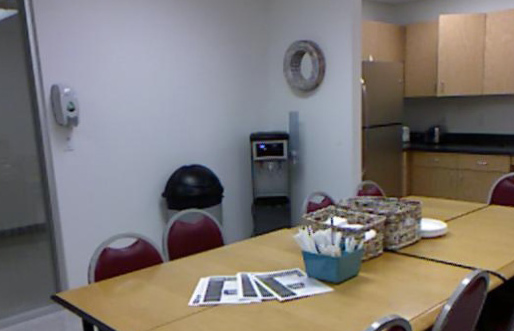}&
		\includegraphics[width=0.49\linewidth]{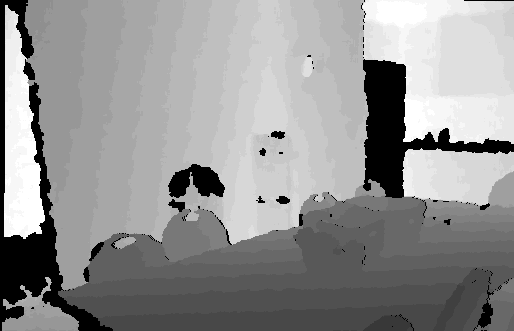}\\	
		\includegraphics[width=0.49\linewidth]{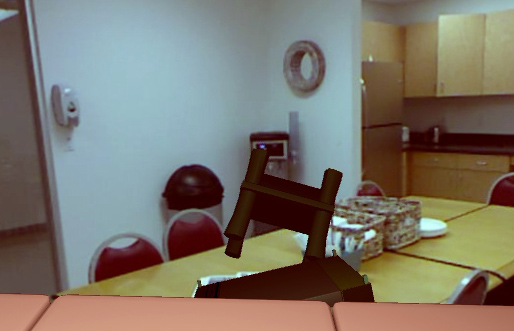}&
		\includegraphics[width=0.49\linewidth]{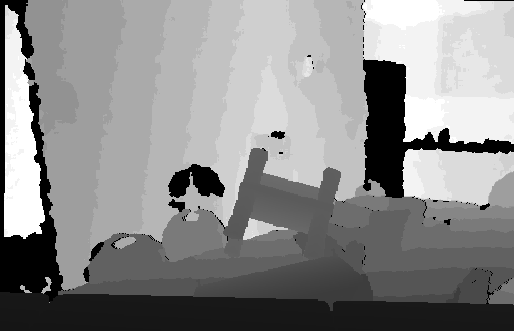}\\	\includegraphics[width=0.49\linewidth]{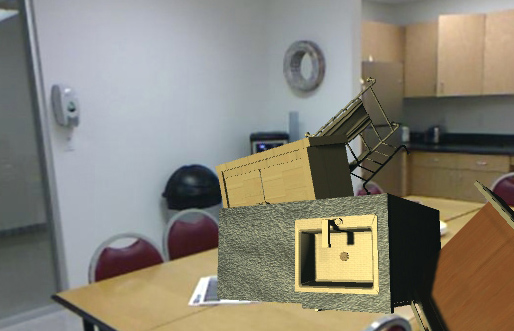}&	
		\includegraphics[width=0.49\linewidth]{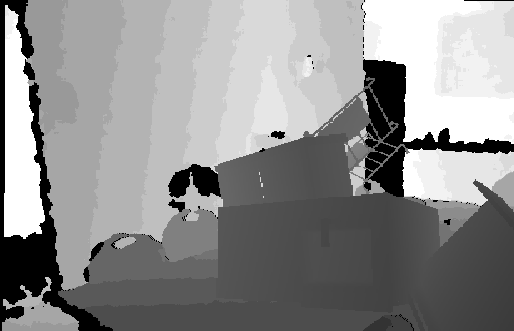}
	\end{tabular}
	\vspace{2.0mm}
	\caption{RGB-D image pairs: the original one (top) and enriched with virtual objects (the rest).}
	\label{fig:enrichment8}
\end{figure*}

Examples of RGB-D NYU-v2 image enrichment with virtual objects are presented in the  \cref{fig:enrichment1} -- \ref{fig:enrichment8}.

\end{document}